\documentclass[journal,comsoc]{IEEEtran}
\usepackage{amsmath,amsthm,amssymb,amsfonts}
\usepackage{bm}
\usepackage{bbm}
\usepackage{booktabs}
\usepackage{algorithm}
\usepackage{algcompatible}
\usepackage{array}
\usepackage[caption=false,font=normalsize,labelfont=sf,textfont=sf]{subfig}
\usepackage{soul}
\usepackage{textcomp}
\usepackage{stfloats}
\usepackage{url}
\usepackage{verbatim}
\usepackage{graphicx}

\usepackage{multirow}
\usepackage{float}
\usepackage{makecell}

\usepackage{tikz,xcolor}

\usepackage{hyperref}
\hypersetup{colorlinks=true, allcolors=black, pdfstartview=Fit}

\definecolor{lime}{HTML}{A6CE39}
\DeclareRobustCommand{\orcidicon}{%
	\begin{tikzpicture}
	\draw[lime, fill=lime] (0,0) 
	circle [radius=0.16] 
	node[white] {{\fontfamily{qag}\selectfont \tiny ID}};
	\draw[white, fill=white] (-0.0625,0.095) 
	circle [radius=0.007];
	\end{tikzpicture}
	\hspace{-2mm}
}

\foreach \x in {A, ..., Z}{%
	\expandafter\xdef\csname orcid\x\endcsname{\noexpand\href{https://orcid.org/\csname orcidauthor\x\endcsname}{\noexpand\orcidicon}}
}

\begin{document}
\title{Take Your Pick: Enabling Effective Personalized Federated Learning within Low-dimensional Feature Space}
\author{Guogang~Zhu\orcidA{}, Xuefeng~Liu\orcidB{}, Shaojie~Tang,
		Jianwei~Niu\orcidD{}, ~\IEEEmembership{Senior~Member,~IEEE}, ~Xinghao~Wu\orcidF{}, and ~Jiaxing~Shen\orcidG{}
		\IEEEcompsocitemizethanks{\IEEEcompsocthanksitem G. Zhu, X. Liu, J. Niu, and X. Wu are with the School of Computer Science and Engineering, Beihang University, China.
			E-mail: buaa\_zgg@buaa.edu.cn, liu\_xuefeng@buaa.edu.cn, niujianwei@buaa.edu.cn, wuxinghao@buaa.edu.cn.
			\IEEEcompsocthanksitem S. Tang is with the Naveen Jindal School of Management, The University of Texas at Dallas. E-mail: tangshaojie@gmail.com. 
            \IEEEcompsocthanksitem J. Shen is with the Department of Computing and Decision Sciences, Lingnan University. E-mail: jiaxingshen@LN.edu.hk.
			\IEEEcompsocthanksitem Corresponding author: Xuefeng Liu.
	}} 
\markboth{Journal of \LaTeX\ Class Files,~Vol.~14, No.~8, August~2021}%
{Shell \MakeLowercase{\textit{et al.}}: A Sample Article Using IEEEtran.cls for IEEE Journals}

\maketitle
\begin{abstract}
Personalized federated learning (PFL) is a popular framework that allows clients to have different models to address application scenarios where clients' data are in different domains. 
The typical model of a client in PFL features a global encoder trained by all clients to extract universal features from the raw data and personalized layers (e.g., a classifier) trained using the client's local data. 
Nonetheless, due to the differences between the data distributions of different clients (aka, domain gaps), the universal features produced by the global encoder largely encompass numerous components irrelevant to a certain client's local task. 
Some recent PFL methods address the above problem by personalizing specific parameters within the encoder. 
However, these methods encounter substantial
challenges attributed to the high dimensionality
and non-linearity of neural network parameter space. 
In contrast, the feature space exhibits a lower dimensionality, providing greater intuitiveness and interpretability as compared to the parameter space.
To this end, we propose a novel PFL framework named FedPick.
FedPick achieves PFL in the low-dimensional feature space by selecting task-relevant features adaptively for each client from the features generated by the global encoder based on its local data distribution. 
It presents a more accessible and interpretable implementation of PFL compared to those methods working in the parameter space.
Extensive experimental results show that FedPick could effectively select task-relevant features for each client and improve model performance in cross-domain FL.

\end{abstract}


\section{Introduction}
\IEEEPARstart{I}{n} recent years, the rapid progress of big data has significantly accelerated the unprecedented growth of deep learning. 
Nonetheless, in real-world scenarios, the data typically originate from geographically dispersed clients, such as mobile phones or wireless sensors \cite{wang2020towards,duan2020self}. 
Due to concerns regarding privacy or communication limitations, centralizing these scattered data for model training is commonly infeasible. 
To address the above challenges, federated learning (FL) emerges as a promising solution that enables multiple clients to collaboratively train the model without sharing their raw data. 
Currently, FL has shown broad prospects for applications in various fields such as mobile edge computing \cite{wu2020accelerating,lu2023auction}, healthcare \cite{chen2020fedhealth,dayan2021federated,yan2023label}, and finance \cite{long2020federated,byrd2020differentially}. However, in practical applications, the data distributions across clients are commonly heterogeneous, which brings significant performance degradation to the FL model \cite{li2020federatedchallenges,kairouz2021advances}. 

Cross-domain FL, where the raw data on different clients come from various domains, is a prevalent source of statistical heterogeneity that appears in practical FL applications. In cross-domain FL, the raw data on different clients are distributed across various spaces, namely $\mathcal{X}_1 \neq \mathcal{X}_2 \neq \cdots \neq \mathcal{X}_N$. However, it is assumed that the labels on different clients share the same space, namely $\mathcal{Y}_1=\mathcal{Y}_2=\cdots=\mathcal{Y}_N$.
For instance, when multiple hospitals collaborate to train a deep model for detecting 
pneumonia (e.g., COVID-19) \cite{dou2021federated}, the diagnostic images (e.g., CT or MRI) from these hospitals can exhibit significant variations due to differences in sensor parameters, scanning protocols, and subject populations \cite{zhang2020collaborative,guan2021domain,wang2022embracing}. Another cross-domain example is autonomous driving \cite{musat2021multi,xiong2023neural}, where the data acquired from different automobiles encompass a wide range of weather conditions, lighting conditions, and geographical locations. The above cross-domain cases introduce domain gaps across clients, leading to a so-called feature shift \cite{li2021fedbn} in the feature space. The feature shift can subsequently degrade the model performance of standard FL methods, such as FedAvg \cite{mcmahan2017communication}. 

Personalized FL (PFL) \cite{mills2021multi,jin2022personalized,tan2022towards} is a widely known means of mitigating the performance degradation in cross-domain FL. 
The fundamental concept behind PFL involves training a personalized model for each client that can adapt to this client's own data distribution with the collaborative assistance of other clients.
Currently, most PFL methods necessitate the sharing of a global encoder among all clients to extract high-level semantics from raw data, while personalizing other components of the model, such as the classifier \cite{FedPer, collins2021exploiting}, to adapt the models to diverse domains. 
It is worth noting that the concept of employing a shared encoder across diverse domains stems from centralized cross-domain learning \cite{cai2019multi}. 
The rationale behind this approach lies in the belief that the shallow parameters in the encoder are less sensitive to data heterogeneity and thus can be applied to various domains \cite{cai2019multi}. 

However, due to the domain gaps across clients in cross-domain FL, the global encoder tends to extract universal features that are applicable for different domains simultaneously. These universal features often encompass numerous components that are irrelevant to the local task of a certain client, thereby potentially impairing the performance of the FL model.
Although some recent PFL methods can tackle the above issue by personalizing specific parameters within the encoder \cite{li2021fedbn,liang2020think,andreux2020siloed,sun2021partialfed}, it is important to note that these methods are commonly conducted within the parameter space. 
Given the complex nature of neural networks, characterized by high dimensionality and non-linearity, identifying an appropriate personalization strategy within the parameter space often proves challenging.
Therefore, we wonder whether it is possible to implement PFL in a low-dimensional feature space, which would offer a more straightforward and interpretable alternative to PFL methods conducted in the parameter space.

\begin{figure}[h]
	\centering
	\includegraphics[width=3.5in]{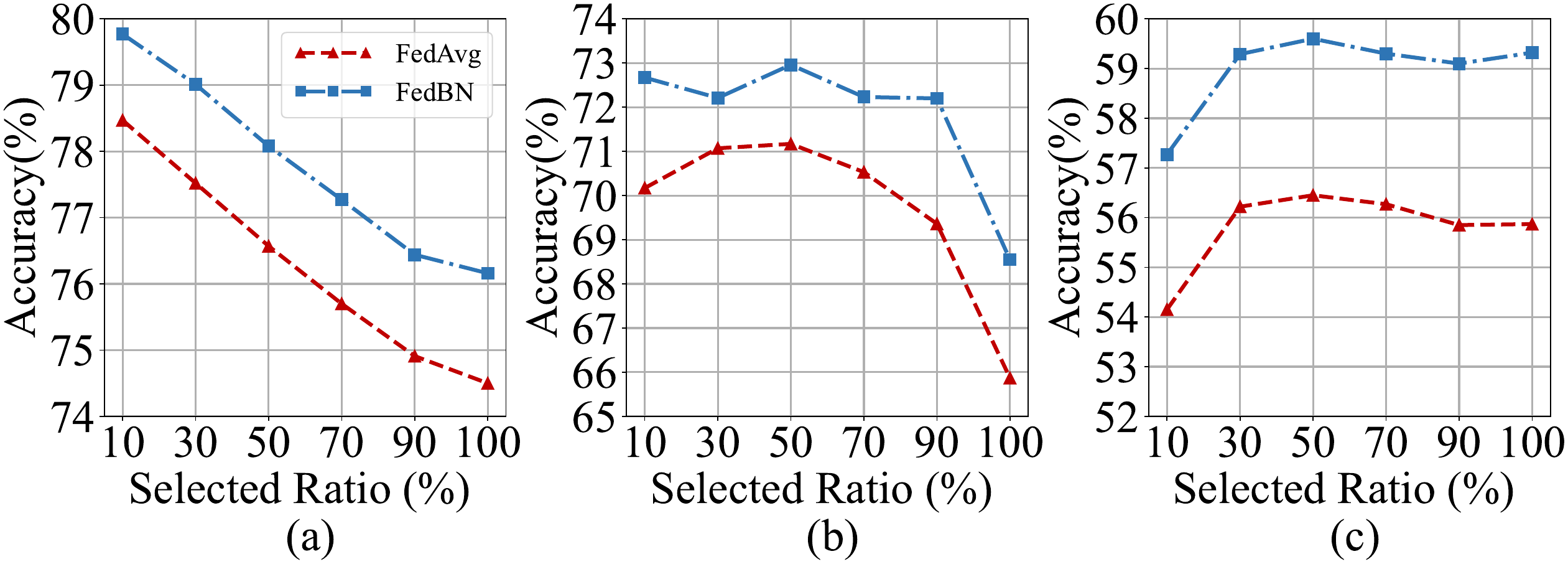}
	\caption{Test accuracy when different 
  feature subsets are selected to retrain a classifier, (a) Digits-Five, (b) Office-Caltech-10, (c) DomainNet.}
        \label{motivation-plot}
\end{figure}

We conduct several experiment to answer the above question. In these experiments, we employ the Fisher Score \cite{geng2020does} to assess the importance of the features.
The Fisher Score, which captures both intra-class consistency and inter-class discrimination, assigns higher scores to components of greater importance in features.
We first arrange the features in descending order based on their Fisher Score, and then select a predetermined proportion of top-ranked features to retrain a classifier to probe the quality of features \cite{chen2020simple,he2022masked,wang2023does}.
Fig. \ref{motivation-plot} illustrates the test accuracy on several commonly used cross-domain datasets using various feature subsets to retrain the classifier.
The highest accuracy occurs when a feature subset with high Fisher Score is selected, surpassing even the accuracy obtained when utilizing all available features.
The above results indicate that even using a simple metric to conduct PFL in the low-dimensional feature space, it is still feasible to achieve relatively good performance.  

However, the above vanilla feature selection approach ignores the correlations among different dimensions of features.
In this paper, we propose FedPick, an approach that can adaptively select a subset of task-relevant features for each client. 
FedPick evaluates the overall quality of the feature subset rather than individually assessing its components, thereby alleviating the drawback of the vanilla feature selection approach.
In FedPick, the global encoder and classifier are retained to extract universal features from raw data. 
It incorporates a personalized feature selection module (PFSM) for each client, enabling the selection of task-relevant features from the universal features. 
Since feature selection is a discretized operation that cannot be directly optimized by commonly used gradient-based algorithms. 
PFSM leverages Gumbel-Sigmoid \cite{jang2016categorical} reparameterization to make feature selection differentiable.
Consequently, PFSM can be optimized simultaneously with the backbone model in an end-to-end manner by local data on individual client. 
Once training is completed, the PFSM can directly output the subset of task-relevant features given the universal features produced by the global encoder.

We conduct comprehensive experiments on multiple commonly used cross domain datasets. The experimental results demonstrate that FedPick can effectively select task-relevant features for each client and subsequently enhance the performance of the FL model. Our contributions of this paper are summarized as follows:
\begin{enumerate}
\item[$\bullet$] We unveil a critical limitation in cross-domain FL, wherein the features generated by the global encoder are frequently redundant and cannot be directly adapted to the local task due to the domain gaps across clients.  
\item[$\bullet$] We propose FedPick, a cross-domain FL method that empowers individual clients to adaptively select task-relevant features based on their local data distribution, thereby enhancing the performance of the FL model.
\item[$\bullet$] We conduct comprehensive experiments to validate the effectiveness of FedPick. The results demonstrate that FedPick significantly improves the model's performance in cross-domain FL scenarios.
\end{enumerate}
\section{Related Work}
\textbf{Statistical Heterogeneity in Federated Learning.} In recent years, FL \cite{mcmahan2017communication,li2020federatedchallenges,kairouz2021advances} has emerged as a promising machine learning paradigm that enables model training without sharing the raw data on local clients. 
In a conventional setting of FL, there is a central server coordinates multiple distributed clients for model training. 
The training procedure involves iterative local training on the clients and model aggregation on the server.
Nonetheless, practical applications often exhibit significant statistical heterogeneity across clients \cite{li2020federatedchallenges,kairouz2021advances}.
Such statistical heterogeneity leads to divergence among locally trained models and subsequently hampers the performance of the aggregated FL model \cite{zhao2018federated}. 
Statistical heterogeneity can manifest in various forms, with cross-domain FL \cite{li2021fedbn} being one of the most common scenarios. 
Cross-domain FL refers to the situation where the distribution of raw input data varies across different clients, which is commonly observed in practical applications such as autonomous driving \cite{musat2021multi,xiong2023neural}, video surveillance \cite{bahnsen2018rain,kenk2020dawn,leroux2022multi}, and medical imaging \cite{zhang2020collaborative,guan2021domain,wang2022embracing}.
Nowadays, several methods are proposed to address the performance degradation in cross-domain FL \cite{li2021fedbn,sun2021partialfed,zhu2022aligning,luo2022disentangled,tan2022federated}. 
Among the aforementioned studies, PFL has effectively showcased its capability to facilitate model adaptation to the local distribution. The subsequent paragraph provides a comprehensive description of PFL methods.

\textbf{Personalized Federated Learning.} 
The primary objective of PFL is to train personalized models for individual clients by leveraging the collaborative efforts of other clients, enabling them to better align with their respective local data distributions. 
FPE \cite{wang2019federate} directly utilizes local data to fine-tune the global model, thereby enhancing its ability to adapt to the local data distribution. 
Meta-learning is integrated into PFL to explore an effective initial model capable of achieving high performance on local clients following a limited number of updates \cite{jiang2019improving,fallah2020personalized}. Parameter decoupling enables PFL by separating personalized parameters from the global model. FedPer \cite{FedPer} and FedRep \cite{collins2021exploiting} both share shallow parameters (e.g., the encoder) while personalizing deep parameters (e.g., the classifier). Nevertheless, in these methods, the universal features produced by the global encoder commonly exhibit limited adaptability to local tasks. Therefore, recent studies try to personalize specific parameters within the encoder to extract features that align local data distribution. LG-FedAvg \cite{liang2020think} employs a contrasting approach to FedPer and FedRep, which establishes local representations and a global head over them. FedBN \cite{li2021fedbn} and SiloBN \cite{andreux2020siloed} address the domain shift in cross-domain FL by localizing BN layers while sharing other parameters. PartialFed \cite{sun2021partialfed} adaptively loads partial rather than entire global parameters at the initialization of local training in cross-domain FL. 
However, the aforementioned studies primarily concentrate on integrating global and local knowledge within the parameter space, where determining an appropriate personalization strategy becomes challenging due to the complexities of high dimensionality and non-linearity. In this paper, we realize PFL within the low-dimensional feature space, which offers the benefits of simplified implementation and enhanced interpretability compared with aforementioned PFL methods that operate in the parameter space.

\section{Problem Formulation of Cross-domain FL}\label{problem_formulation}
In this section, we outline the problem formulation of cross-domain FL. The key notations utilized in this paper are listed in Table \ref{notations}.

FL commonly involves a central server that coordinates $N$ distributed clients to perform model training without sharing their private data. Suppose that each client consists of $M_i$ samples that are generated from $\mathcal{D}_i$, which are denoted as $(\bm{x}_i^j, \bm{y}_i^j), j=1,2,\dots,M_i$. Specifically, $\bm{x}_i^j \in \mathcal{X}_i \subseteq \mathbb{R}^n$ denotes the raw input, and $\bm{y}_i^j \in \mathcal{Y}_i$ denotes the corresponding label. In cross-domain FL, the raw data on different clients come from various domains, that is, $\mathcal{X}_1 \neq \mathcal{X}_2 \neq \cdots \neq \mathcal{X}_N$. However, the labels on different clients are assumed to be uniformly distributed in the same space, that is, $\mathcal{Y}_1=\mathcal{Y}_2=\cdots=\mathcal{Y}_N$. Moreover, we only consider the classification task in this paper, so $\mathcal{Y}_1=\mathcal{Y}_2=\cdots=\mathcal{Y}_N = \{1, 2, \dots, C\}$, where $C$ is the total number of classes.

We follow the training paradigm of PFL, whose core idea is to train a personalized model $\bm{\theta}_i$ for each client that can adapt to the client's data distribution, as shown below:
\begin{equation}
\label{PFL}
\underset{\bm{\theta_1}, \bm{\theta_2}, \dots, \bm{\theta_N}}{\text{argmin}} \frac{1}{N} \sum_{i=1}^{N}\mathcal{L}_i(\bm{\theta_i};\mathcal{D}_i),
\end{equation}
where $\mathcal{L}_i$ denotes the expected risk on client $i$. In practice, the expected risk $\mathcal{L}_i$ is often inaccessible due to the unavailability of the underlying data distribution $\mathcal{D}_i$. Therefore,  the empirical risk $\hat{\mathcal{L}}_i(\bm{\theta}_i)$ on empirical data distribution $\hat{\mathcal{D}}_i$ is frequently employed as an approximation for the expected risk $\mathcal{L}_i$, which is formulated as follows:
\begin{equation}
\label{empirical-risk}
\mathcal{L}_i(\bm{\theta}_i) \approx \hat{\mathcal{L}}_i(\bm{\theta}_i) = \frac{1}{M_i} \sum_{j=1}^{M_i} \ell (\bm{y}_i^j, \hat{\bm{y}}_i^j),
\end{equation}
where $\hat{\bm{y}}_i^j = f_i(\bm{x}_i^j;\bm{\theta}_i)$ is the predicted label, and $\ell:\mathcal{Y} \times \mathcal{Y} \rightarrow \mathbb{R}$ denotes the loss function that measures the prediction error.

Generally, $\bm{\theta}_i$ can be decomposed into two parts: an encoder $\bm{\phi}_i$ (typically composed of stacked convolutional layers) and a classifier $\bm{h}_i$ (typically composed of one or more fully connected layers). The encoder $\bm{\phi}_i: \mathbb{R}^n \rightarrow \mathbb{R}^k$ maps raw inputs $\mathcal{X}_i \subseteq \mathbb{R}^n$ to a lower-dimensional feature space $\mathcal{Z}_i \subseteq \mathbb{R}^k$, which is denoted as $\bm{z}_i = \bm{\phi}_i(\bm{x}_i)$ and typically $k \ll n$ in practice. The classifier $\bm{h}_i: \mathbb{R}^k \rightarrow \mathcal{Y}$ gives the final prediction $\hat{\bm{y}}_i$ based on the feature $\bm{z}_i$, which is denoted as $\hat{\bm{y}}_i=\bm{h}_i(\bm{z}_i)$. As discussed in following subsections, we empirically observe that $\bm{z}_i$ is redundant for $\bm{h}_i$ to accomplish the local task in cross-domain FL. Specifically, some components in $\bm{z}_i$ are irrelevant or even harmful to the local task on each client. Therefore, we propose FedPick, whose main purpose is to select a subset of task-relevant features from $\bm{z}_i$ to adapt to the local data distribution of each client. 

\begin{table}[h]
    \centering
    \renewcommand{\arraystretch}{1.3}
    \caption{List of key Notations.}
    \label{notations}
    \begin{tabular}{c | c}
    \bottomrule
         Symbol & Description \\  \hline
         \multicolumn{2}{c}{\textbf{Federated Learning System}} \\ \hline
         $N$ & number of clients participating the FL training \\ \hline
         $M_i$ & number of samples on client $i$ \\ \hline
         $C$ & total number of classes of samples \\ \hline
         $(\bm{x}_i^j, \bm{y}_i^j)$ & the $j_{th}$ training sample  on client $i$ \\ \hline

         \multicolumn{2}{c}{\textbf{Model Architecture}} \\ \hline
         $\bm{\theta}_i$ & personalized model on client $i$ \\ \hline
         $\bm{\phi}^g$ & global encoder \\ \hline
         $\bm{h}^g / \bm{h}^p_i / \bm{h}^u_i$ & classifier for global / task-relevant / irrelevant features \\ \hline
         
         \multicolumn{2}{c}{\textbf{Feature Selection}} \\ \hline
         $S_b / S_w$ & inter-class / intra-class variance of features \\ \hline
         $G', G''$ & noise for Gumbel sampling \\ \hline
         $\bm{z}_{i}^{l}$ & unbounded logits \\ \hline
         $\bm{z}_{i}^{g} / \bm{z}_{i}^{p} / \bm{z}_{i}^{u}$ & global / task-relevant / irrelevant features \\ \hline
         $\bm{m}_i^s / \bm{m}_i$ & soft / hard feature mask\\ \hline
         $\hat{\bm{y}}_i^g / \hat{\bm{y}}_i^p / \hat{\bm{y}}_i^u$ & predictions of global / task-relevant / irrelevant features \\ \hline
         
         \multicolumn{2}{c}{\textbf{Loss Function}} \\ \hline
         $\mathcal{L}_{ent}$ &  entropy of predictions from task irrelevant features \\ \hline
         $\mathcal{L}_{lce}$ &  cross entropy of predictions from personalized features \\ \hline
         $\mathcal{L}_{dis}$ &  distillation between personalized and global predictions \\ \hline
         $\mathcal{L}_{gce}$ &  cross entropy of predictions from global features \\
    \toprule
    \end{tabular}
\end{table}
 
\section{Motivation of Feature Selection in Cross-domain FL}\label{handcraft}
In this section, we discuss the motivation of personalized feature selection in cross-domain FL. At first, we illustrate the observation that the universal features generated by the global encoder in FL often exhibit a higher degree of redundancy compared to the features generated from models trained through centralized learning (CL). Then we demonstrate that the performance of the model can be improved by selecting an appropriate subset of features tailored to the local task.
\subsubsection{Feature Redundancy}
We employ the sparsity ratio as quantitative metric to assess the feature redundancy. 
The sparsity ratio is defined as the percentage of components that are closed to zero in the features. 
We apply a threshold value $\varepsilon$ to each individual component of $L_2$-normalized features to determine whether it is closed to zero. 
The formulation for computing the sparsity ratio $S (\bm{z}, \varepsilon)$ is as follows:

\begin{equation}
    \label{sparsity}
    S (\bm{z}, \varepsilon) = \frac{||M(\overline{\bm{z}}, \varepsilon)||_1}{|\bm{z}|},
\end{equation}
\begin{equation}
    \label{mask}
    M(\overline{\bm{z}}, \varepsilon) [ i ]  = \left\{
    \begin{aligned}
        & 1, \quad if \ \overline{\bm{z}}^i \leq \varepsilon \\
        & 0, \quad if \ \overline{\bm{z}}^i > \varepsilon,
    \end{aligned}
    \right.
\end{equation}
where $\varepsilon$ is the threshold value, $\overline{\bm{z}}$ is the $L_2$-normalized features, $||\cdot||_1$ is the $L_1$ norm of vector, $|\bm{z}|$ is the dimension of $\bm{z}$.

We conduct experiments to measure the sparsity ratio of features generated by different methods on Digits-Five (a commonly used cross-domain dataset), including FedAvg \cite{mcmahan2017communication}, FedBN \cite{li2021fedbn} and SingleSet (training an individual model for each domain). 
To mitigate mutual interference between multiple domains, we train a separate model for each domain (i.e., SingleSet) for CL training paradigm, instead of training a model using the data from all domains. 
The threshold value $\varepsilon$ in Eq. (\ref{sparsity}) and Eq. (\ref{mask}) is set to $10^{-5}$.

\begin{figure}[h]
	\centering
	\includegraphics[width=3in]{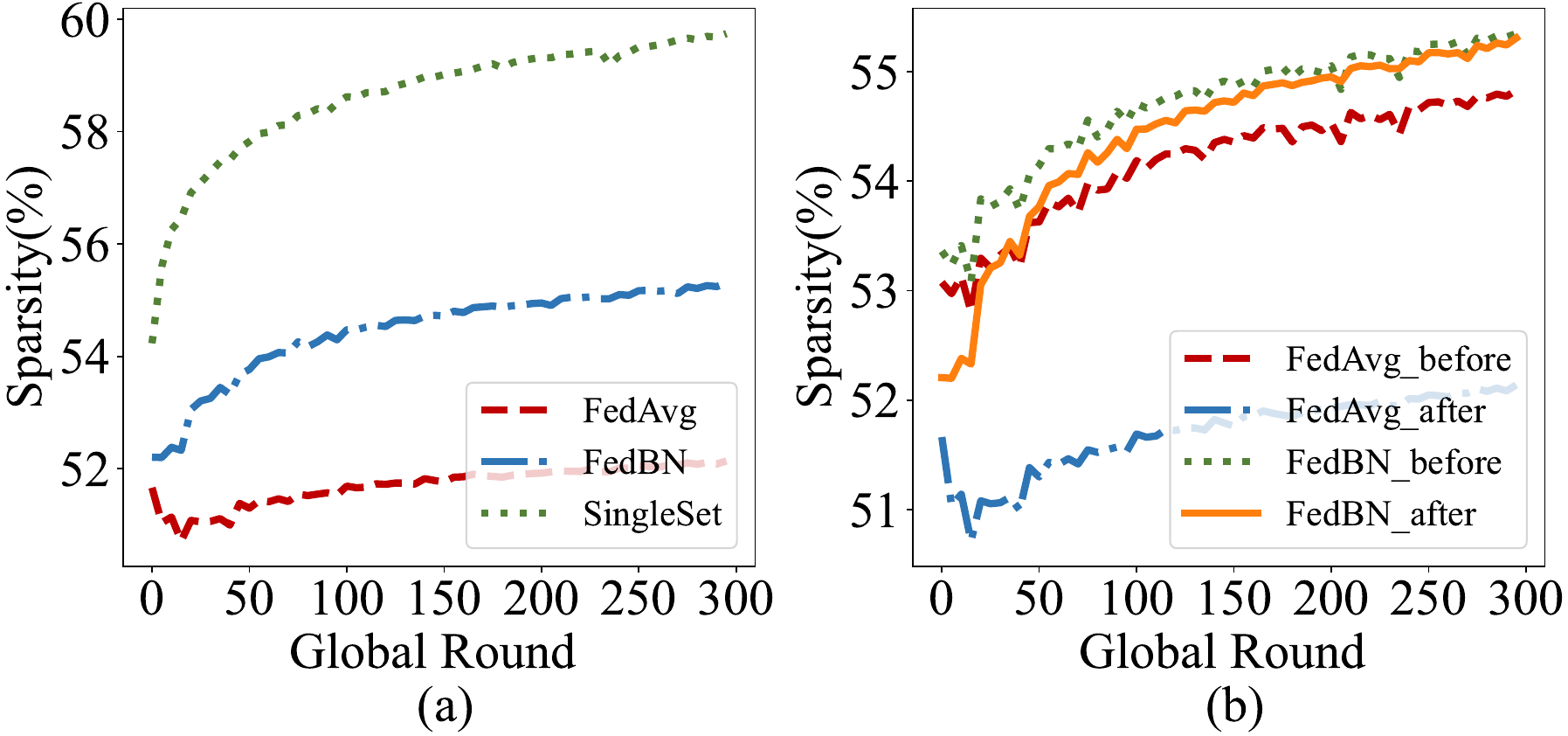}
	\caption{Analysis of feature redundancy, (a) sparsity ratio of different methods, (b) sparsity ratio before and after model aggregation.}
        \label{sparsity_all}
\end{figure}

Fig. \ref{sparsity_all} (a) presents the average sparsity ratio of features generated by different methods on various domains in Digits-Five. 
As expected, the features exhibit increased sparsity as the training progresses. This observation can be attributed to the fact that the ground truth is represented by an one-hot vector, which inherently possesses extreme sparsity. Consequently, the features near the classifier should strive for maximum sparsity to minimize the training loss.
Moreover, it can be observed that the sparsity ratio of FL methods is notably lower than that of SingleSet.
These findings suggest that the features generated by FL methods exhibit greater redundancy compared to those produced by the CL method, which deviates from the training objective.
One plausible explanation for this disparity is that FL methods aim to generate universal features that are versatile enough to be shared across various domains, whereas SingleSet models only need to meet the requirements of the local data distribution within that specific domain. 

Additionally, we demonstrate that the increase in feature redundancy is a consequence of model aggregation. Fig. \ref{sparsity_all} (b) presents a comparative analysis of the sparsity ratios of features both before and after the model aggregation. The results clearly indicate a substantial rise in redundancy among the features generated by the model after aggregation, in contrast to those originating from the models prior to aggregation.
\subsubsection{Model Performance with Different Feature Subsets}
\begin{figure}[h]
	\centering
	\includegraphics[width=2.8in]{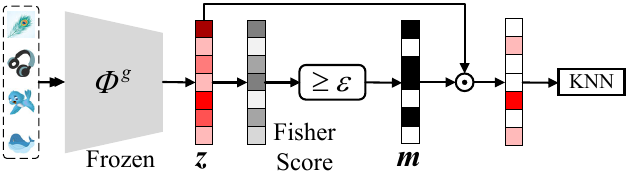}
	\caption{Procedure of vanilla feature selection.}
        \label{vanilla}
\end{figure}

Inspired by the above observation, we posit that improving feature redundancy can enhancing the generalization ability of the learned model to all clients. 
Nonetheless, this enhancement comes at the cost of diminishing the model's generalization  ability when confronted with the local distribution of individual clients, i.e., the personalization ability. 
Therefore, we pose the following question: \textbf{\textit{Can model performance be improved by selecting an approximated feature subset from the universal features?}}
To answer the above question, we utilize Fisher Score \cite{gu2012generalized} to assess the feature importance and perform a vanilla feature selection approach, as shown in Fig. \ref{vanilla}. Fisher Score is a widely known metric that can evaluate the importance of an individual component in features, which is defined as follows:
\begin{equation}
    \label{fisher_score}
    F_i = \frac{S_b}{S_w},
\end{equation}
where $S_b$ is the inter-class variance, $S_w$ is intra-class variance. The inter-class variance $S_b$ is defined as:
\begin{equation}
    \label{inter-class}
    S_b = \sum_{j=1}^{C}m_j(\mu_{ij} - \mu_i)^2,
\end{equation}
where $m_j$ is the number of samples in class $j$, $\mu_{i}$ is the mean of feature $z_i$, $\mu_{ij}$ is the mean of feature $z_i$ for samples in class $j$.
The intra-class variance $S_w$ is defined as: 
\begin{equation}
     \label{intra-class}
     S_w = \sum_{j=1}^{C}m_j\sigma_{ij}^2,
\end{equation}
where $m_j$ is the number of samples in class $j$, $\sigma_{ij}$ is the variance of feature $z_i$ for samples in class $j$.
A higher Fisher Score is achieved when the component exhibits greater similarity within the same class and dissimilarity across different classes, which implies that this components is more discriminative and is more relevant to the local task.

In the vanilla feature selection, we first pre-train a model using existed FL methods until it converges. Then we calculate the Fisher Score of features on the training dataset and constitute the feature mask $\bm{m}$ based on the Fisher Score. $\bm{m}$ is a binary vector, wherein positions exceeding a specified threshold based on the Fisher Score are assigned a value of 1, while the remaining positions are assigned a value of 0. In the experiments, the threshold is defined as the percentile obtained from the sorted Fisher Score vector based on the selected ratio. At last, we freeze the encoder and implements a probe \cite{chen2020simple,he2022masked,wang2023does} on the masked features, denoted as $\bm{z} \odot \bm{m}$, to quantitatively assess their quality. Specifically, we discard the previously trained classifier and employ the masked features to retrain a new classifier.

\begin{figure*}[!ht]
	\centering
	\includegraphics[width=7in]{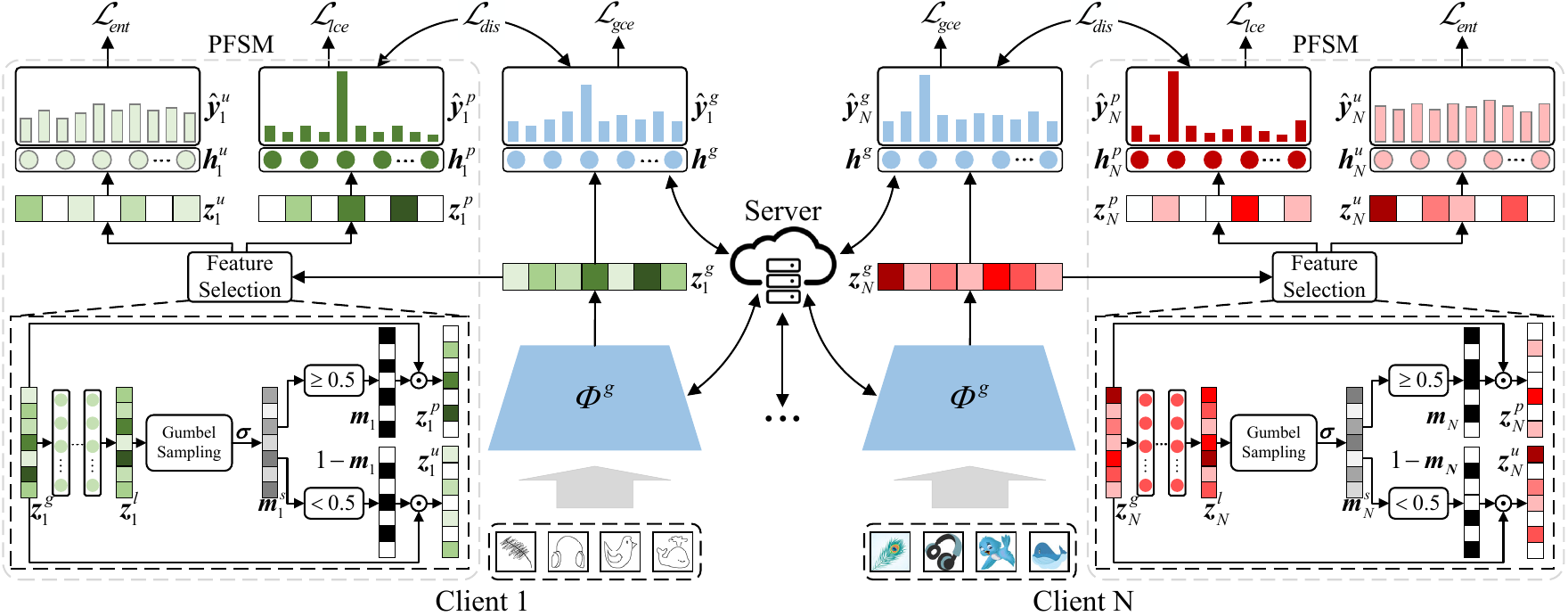}
	\caption{Framework of FedPick. It mainly consists of the following steps: (1) The raw data are passed through a global encoder to generate universal features. (2) The universal features are subsequently fed into a PFSM to select personalized task-relevant features. (3) Both the universal features and the personalized features are then utilized as inputs for a global classifier and a personalized classifier, respectively, to generate the prediction.}
        \label{framework}
\end{figure*}

We employ FedAvg and FedBN to pre-train the FL models. During the feature selection, the component indexes in features are sorted in descending order based on the Fisher Score. 
The selection ratio ranges from $10\% - 100 \%$.
We employ a K Nearest Neighbor (KNN) classifier (with $K=10$) to probe the feature subsets.  
Fig. \ref{motivation-plot} presents the test accuracy achieved with different feature subsets. 
It can be observed that by adaptively choosing task-relevant features (those with high Fisher Scores) for the downstream task, the model can outperform the performance achieved using all available features.

However, the aforementioned vanilla feature selection method encounters several challenges. Firstly, although the Fisher Score has been proved to be effective for the classification task, designing a practical and applicable feature evaluation metric remains a challenge in real-world scenarios. Secondly, the above approach evaluates features independently, thereby overlooking the interrelationships among different dimensions in the feature. Lastly, determining the optimal proportion of selected components poses a challenging problem.
To address these issues, we further propose a novel approach called FedPick. FedPick facilitates the automatic selection of task-relevant features based on the local data distribution of each client. By leveraging this method, we aim to overcome the limitations of vanilla feature selection method and enhance the performance of FL models.

\section{Framework of FedPick}
In this section, we first provide an overview of FedPick. Second, we introduce personalized feature selection module (PFSM), the core part of FedPick. Then, we discuss the knowledge transfer mechanism between global and personalized features. At last, we provide the training procedure of FedPick.
\subsection{Framework Overview}
The framework of FedPick is depicted in Fig. \ref{framework}, which mainly consists of three steps. First, the raw data are passed through a global encoder to generate universal features. Then, the universal features are fed into a PFSM for feature selection, separating the features to a task-relevant feature subset and a task-irrelevant feature subset. At last, the universal features and personalized task-relevant features (for brevity, we sometimes refer to these features as ‘personalized features’ or ‘task-relevant features’ in the subsequent sections) are passed into a global and a personalized classifier for prediction, respectively. During inference, the global and personalized predictions are integrated to derive the final prediction. For simplification, we omit the index of clients and samples in this section.

\subsection{Personalized Feature Selection Module}
In FedPick, the encoder $\bm{\phi}^g$ is shared across multiple clients to extract universal features. As previously mentioned, the universal features generated by the global encoder exhibit a higher level of generalization for a wider distribution, offering potential advantages in subsequent feature selection processes. To preserve the universality of global features, a global classifier $\bm{h}^g$ (also shared across clients) is maintained within FedPick. The features generated by the global encoder are denoted as $\bm{z}^g$, and the corresponding predictions are represented as $\hat{\bm{y}}^g = \bm{h}^g (\bm{z}^g)$. The features generated by the global encoder are required to possess sufficient discriminative power to successfully carry out the classification task. This objective is achieved by cross entropy loss, as demonstrated in Eq. (\ref{loss_CE_global}).
\begin{equation}
    \label{loss_CE_global}
    L_{gce} = \sum_{c=1}^{C} y_c \log (\hat{y}^g_c)
\end{equation}

After generating universal features from the global encoder, FedPick introduces a PFSM for each client to adaptively select each client's task-relevant features based on its local data distribution. In PFSM, the global features $\bm{z}^g$ are first fed into a FC network, denoted as $\bm{\varphi}$, to generate the unbounded logits $\bm{z}^l$. Subsequently, the Gumbel-Sigmoid reparameterization technique, as described in \cite{geng2020does}, is employed to generate a soft mask denoted as $\bm{m}^{s}$. Specifically, the calculation of the $i_{th}$ component's corresponding mask $\bm{z}^l$ is as follows:
\begin{equation}
\label{gumbel}
m_i^s =\sigma((z^l_i + G' -G'') / \tau) = \frac{e^{(z^l_i + G') / \tau}}{e^{(z^l_i + G')/ \tau} + e^{G'' / \tau}},  
\end{equation}
where $G'$ and $G''$ are two independent Gumbel noises sampled from uniform distribution $U[0,1]$, $\tau \in [0, +\infty]$ is the temperature scale that controls the distribution tendency of sampling, $\sigma(\cdot)$ is the sigmoid function.
It should note that the noises $G'$ and $G''$ are activated during training to facilitate the exploration of various feature masks. 
However, during inference, these noises are deactivated to ensure consistent and reliable results.
Similar to vanilla feature selection approach, the soft mask $\bm{m}^s$ is then discretized into a binary vector $\bm{m}$ by a threshold $\varepsilon$, that is, $m_i$ is set to 1 if $m^s_i \geq \varepsilon$ else 0, as depicted in Eq. (\ref{mask}). In FedPick, unless otherwise specified, the value of $\varepsilon$ is set to 0.5.

However, the previously mentioned hard masking operation is discrete, which cannot be directly optimized by commonly used gradient-based algorithms. To make the hard masking differentiable,  FedPick adopts sigmoid during the backward process and hard masking during the forward process \cite{bengio2013estimating}. This design enables simultaneous optimization of PFSM with the backbone model using the training data and can be easily implemented within popular deep learning frameworks such as PyTorch \cite{paszke2019pytorch}.

PFSM outputs task-relevant features $\bm{z}^p$ and task-irrelevant features $\bm{z}^u$ by Hadamard product between the $\bm{z}^g$ and $\bm{m}$, $1-\bm{m}$, respectively, as shown in Eq. (\ref{relevant_feature}).
\begin{equation}
\label{relevant_feature}
\bm{z}^p = \bm{z}^g \odot \bm{m}, \quad
\bm{z}^u = \bm{z}^g \odot (1 - \bm{m}).
\end{equation}

Similar to $\bm{z}^g$, the task-relevant features $\bm{z}^p$ also shall be discriminative enough to accomplish the local task. Therefore, PFSM implements a personalized classifier $\bm{h}^p$ for $\bm{z}^p$ and enforces it to accomplish the classification task by minimizing the cross entropy loss, as shown in Eq. (\ref{loss_CE_local}).
\begin{equation}
    \label{loss_CE_local}
    L_{lce} = \sum_{c=1}^{C} y_c \log (\hat{y}^p_c)
\end{equation}

In contrast, the task-irrelevant features $\bm{z}^u$ are expected to exhibit lower levels of discriminative capability towards the local task. As a result, the PFSM employs a personalized classifier $\bm{h}^u$ to $\bm{z}^u$ and encourages its prediction to be uncertain by maximizing its prediction, as illustrated in Eq. (\ref{loss_entropy}).
\begin{equation}
    \label{loss_entropy}
    L_{ent} = \sum_{c=1}^{C} \hat{y}^u_c \log (\hat{y}^u_c)
\end{equation}

\subsection{Transferring Global and Personalized Knowledge}
In FedPick, the global features and personalized features offer distinct benefits. Global features possess a higher level of generalization on different data distribution, allowing for their transferability across various distributions. However, this high transferability may result in performance degradation when dealing with local data distributions on individual clients. Conversely, personalized features are tailored to local distributions, thereby enhancing performance for specific distributions on local clients (personalization). Nonetheless, these features may lack generalization capabilities when confronted with unseen data distributions, such as test datasets. Hence, it is preferable to combine the advantages of both global and personalized features to enhance the overall model performance. Motivated by knowledge distillation \cite{hinton2015distilling}, FedPick incorporates cyclic distillation between the predictions from global and personalized features to facilitate mutual knowledge transfer. The cyclic distillation of global and personalized features ensures the balance between the model's generalization and personalization abilities across different data distributions. This cyclic distillation is achieved by minimizing the loss function presented in Eq. (\ref{loss_dis}), where $KL()$ denotes the Kullback-Leibler (KL) divergence, which quantifies the dissimilarity between two distributions.
\begin{equation}
    \label{loss_dis}
    L_{dis} = KL(\hat{\bm{y}}^p||\hat{\bm{y}}^g) + KL(\hat{\bm{y}}^g||\hat{\bm{y}}^p)
\end{equation}

To leverage the comprehensive knowledge provided by both global and personalized features, FedPick combines the predictions of the global and personalized classifiers through an ensemble approach, resulting in a more robust prediction. This ensemble process is illustrated in Eq. (\ref{ensemble}). 
\begin{equation}
    \label{ensemble}
    \hat{\bm{y}} = softmax(\hat{\bm{y}}^g + \hat{\bm{y}}^p)
\end{equation}

\subsection{Training Procedure of FedPick}
The total loss adopted by FedPick is shown in Eq. (\ref{loss_total}), where $\lambda_{lce}$, $\lambda_{ent}$, and $\lambda_{dis}$ are hyper-parameters to adjust the effect of different loss terms. During the training, all parameters in the model are optimized by the loss in Eq. (\ref{loss_total}) simultaneously. 
\begin{equation}
    \label{loss_total}
    L_{total} = L_{gce}  + \lambda_{lce} L_{lce} + \lambda_{ent} L_{ent} + \lambda_{dis} L_{dis}
\end{equation}

After training, the global encoder and classifier are uploaded to the server for aggregation, as shown in Eq. (\ref{global-aggregation}), where $\bm{\phi}^g_i$ and $\bm{h}^g_i$ denote the global encoder and classifier updated on client $i$, $\tilde{\bm{\phi}}^g$ and $\tilde{\bm{h}}^g$ the aggregated global encoder and classifier, respectively. $\tilde{\bm{\phi}}^g$ and $\tilde{\bm{h}}^g$ are then broadcast to each client for next round of local updates. 
The parameters in PFSM are localized on each client to select its own task-relevant features. Moreover, motivated by previous studies in \cite{li2021fedbn,andreux2020siloed}, FedPick also keeps BN layers personalized on each client to avoid the performance degradation caused by aggregating them.

\begin{equation}
\label{global-aggregation}
\tilde{\bm{h}}^g = \sum_{i=1}^{N} \frac{M_i}{\sum_{j=1}^{N}M_j}\bm{h}^g_i , \quad \tilde{\bm{\phi}}^g = \sum_{i=1}^{N} \frac{M_i}{\sum_{j=1}^{N}M_j}\bm{\phi}^g_i .
\end{equation}

The entire process of FedPick is summarized in Algorithm \ref{alg-FedPick}. For brevity, we denote the parameters in PFSM as $\bm{\psi}_i$, that is, $\bm{\psi}_i \equiv \{\bm{h}_i^p, \bm{h}_i^u, \bm{\varphi}_i\}$.

\begin{algorithm}[h]
    \caption{FedPick}
    \label{alg-FedPick}
    \begin{algorithmic}[1]
        \STATEx \textbf{Notations:} $T$: global update rounds, $E$: local update epochs, $B$: local minibatch size, $\eta$: learning rate, $\lambda_{lce}$, $\lambda_{ent}$, and $\lambda_{dis}$: hyperparameters used to balance loss terms.
        \STATEx \textbf{Sever Executes:}
        \STATE initialize and broadcast $\bm{\phi}^{g,1}, \bm{h}^{g,1}, \bm{\psi}_1^1, ...,\bm{\psi}_N^1$ to clients
    
        \FOR{$t=1,2,3,\dots,T$}
        
        \FOR {each client $i$ \textbf{in parallel}}
        \STATE $\bm{\phi}_i^{g,t+1}, \bm{h}_i^{g,t+1} \leftarrow \text{ClientUpdate(}i, t, \tilde{\bm{\phi}}^{g,t}, \tilde{\bm{h}}^{g,t}, \bm{\psi}_i^t$)
        \ENDFOR
        
        \STATE $\tilde{\bm{h}}^{g,t+1} = \sum_{i=1}^{N} \frac{M_i}{\sum_{j=1}^{N}M_j}\bm{h}^{g,t}_i$ 
        \STATE $\tilde{\bm{\phi}}^{g,t+1} = \sum_{i=1}^{N} \frac{M_i}{\sum_{j=1}^{N}M_j}\bm{\phi}^{g,t}_i$
        \STATE broadcast $\tilde{\bm{\phi}}^{g,t+1}$, $\tilde{\bm{h}}^{g,t+1}$ to clients
        \ENDFOR
        
        \STATEx \textbf{ClientUpdate($i, t, \bm{\phi}, \bm{h}, \bm{\psi}$):}
        \STATE $\mathcal{B} \leftarrow$ (split local dataset into batches of size B)
        \FOR{$j=1,2,3,\dots, E$}
        \FOR {batch $b \in \mathcal{B}$}
        \STATE $\bm{z}^g = \bm{\phi}(b), \ \{\bm{z}^p, \bm{z}^u \} = \bm{\psi}(\bm{z}^g)$
        \STATE $\hat{\bm{y}}^g = \bm{h}^g(\bm{z}^g), \ \hat{\bm{y}}^p = \bm{h}^p(\bm{z}^p), \ \hat{\bm{y}}^u = \bm{h}^u(\bm{z}^u)$
        \STATE $L_{total} = L_{gce}(\hat{\bm{y}}_g)  + \lambda_{lce} L_{lce}(\hat{\bm{y}}^p)$
        \STATEx $ \qquad \qquad \qquad + \lambda_{ent} L_{ent}(\hat{\bm{y}}^u) + \lambda_{dis} L_{dis}(\hat{\bm{y}}^p,\hat{\bm{y}}^u)$
        \STATE $\{\bm{\phi}, \bm{h}, \bm{\psi}\} \leftarrow \{\bm{\phi}, \bm{h}, \bm{\psi}\} - \eta \bigtriangledown_{\{\bm{\phi}, \bm{h}, \bm{\psi}\}}L_{total}$
        \ENDFOR
        \ENDFOR
        \STATE $\bm{\psi}_i^{t+1} \leftarrow \bm{\psi}$
        \STATE return $\bm{\phi}$, $\bm{h}$ to server
        
    \end{algorithmic}
\end{algorithm}

\section{Theoretical Foundation of FedPick}
FedPick selects appropriate feature subset relevant to the local task from the global features, thereby enhancing their compatibility with the local distribution on each client. Such a feature selection operation can be theoretically supported by Vapnik-Chervonenkis (VC) theory \cite{cristianini2000introduction}. The VC dimension serves as a quantifiable measure of the capacity of a hypothesis class, which represents the set of possible functions that a learning algorithm can learn. Specifically, it establishes an upper bound on the number of training samples that can be perfectly classified by the hypothesis class. When a hypothesis class exhibits a high VC dimension, it indicates a larger capacity, enabling the class to potentially capture a wide range of complex patterns present in the training data. 

As described in Section \ref{problem_formulation}, each client's model consists of an encoder and a classifier. The encoder $\bm{\phi}_i: \mathbb{R}^n \rightarrow \mathbb{R}^k$ is responsible for mapping the raw inputs $\mathcal{X}_i \subseteq \mathbb{R}^n$ to a lower-dimensional feature space $\mathcal{Z}_i \subseteq \mathbb{R}^k$, denoted as $\bm{z}_i = \bm{\phi}_i(\bm{x}_i)$. The classifier $\bm{h}_i: \mathbb{R}^k \rightarrow \mathcal{Y}$ generates the final prediction $\hat{\bm{y}}_i$ based on the extracted feature $\bm{z}_i$, expressed as $\hat{\bm{y}}_i=\bm{h}_i(\bm{z}_i)$. The classifier $\bm{h}_i$ is sampled from the hypothesis space $\mathcal{H}$. When considering only the stage after feature extraction, the optimization objective of the FL system can be reformulated as follows:

\begin{equation}
\label{PFL_classifier}
\underset{\bm{h_1}, \bm{h_2}, \dots, \bm{h_N}}{\text{argmin}} \frac{1}{N} \sum_{i=1}^{N}\mathcal{L}_i(\bm{h_i}; \bm{\phi}_i(\mathcal{D}_i)),
\end{equation}

For client $i$, it is proved by VC dimension theory that with a probability of at least $1 - \delta$, the expected risk $\mathcal{L}_i(\bm{h}_i)$ is upper bounded by: 
\begin{equation}
\mathcal{L}_i(\bm{h}_i) \leq \hat{\mathcal{L}}_i(\bm{h}_i) + \sqrt{\frac{8 \gamma \ln(2M_i) + 8 \ln \frac{4}{\delta}}{M_i}},
\label{vc-dimensional-local}
\end{equation}
where $\hat{\mathcal{L}}_i(\bm{h}_i)$ is the empirical risk, $M_i$ is the number of training samples on client $i$, $\bm{h}_i$ is the classification model distributed in hypothesis space $\mathcal{H}$, $\gamma$ is the VC dimension of $\mathcal{H}$. 

Denoting the global data distribution as $\mathcal{D}_g = \sum_{i=1}^{N} \lambda_i \mathcal{D}_i$, where $\lambda_i = \frac{M_i}{M}$ and $M = \sum_{i=1}^{N} M_i$. We use the notation $d_{\mathcal{H}}(\mathcal{D}_1, \mathcal{D}_2)$ to represent the divergence between two distributions. Based on the global model generalization proposed by previous studies \cite{mohri2019agnostic,mansour2020three}, with a probability of at least $1-\delta$, the risk of client $i$ in the FL system is bounded by:

\begin{equation}
\mathcal{L}_i(\bm{h}_i) \leq \hat{\mathcal{L}}_i(\bm{h}_g) + \sqrt{\frac{8 \gamma \ln(2M) + 8 \ln \frac{4}{\delta}}{M}} + d_{\mathcal{H}}(\mathcal{D}_i, \mathcal{D}_g),
\label{vc-dimensional-single}
\end{equation}
where $\bm{h}_g$ denotes the global classier obtained from $\mathcal{D}_g$.

Combining the global optimization objective in Eq. (\ref{PFL_classifier}) with Eq. (\ref{vc-dimensional-single}) yields the generalization bound of FL system as:

\begin{equation}
\label{vc-dimensional-all}
\begin{split}
    \frac{1}{N} \sum_{i=1}^{N} \underset{\bm{h_i} \in \mathcal{H}}{\text{argmin}} \mathcal{L}_i(\bm{h}_i) 
    & \leq \frac{1}{N} \sum_{i=1}^{N} \hat{\mathcal{L}}_i(\bm{h}_g) + \frac{1}{N} \sum_{i=1}^{N} d_{\mathcal{H}}(\mathcal{D}_i, \mathcal{D}_g) \\
    & + \sqrt{\frac{8 \gamma \ln(2M) + 8 \ln \frac{4}{\delta}}{M}}.
\end{split}
\end{equation}

To enhance the generalization of the model, the primary objective is to reduce the upper bound, which can be achieved by increasing the number of training samples. By comparing Eq. (\ref{vc-dimensional-local}) and Eq. (\ref{vc-dimensional-all}), it can be observed that clients can augment the quantity of training samples by FL, thereby bolstering the generalization capability of the local model in contrast to local training. Another approach to tightening the generalization bound is to reduce the VC dimension $\gamma$. For a linear classifier, $\gamma$ is upper bounded by the dimension of the features \cite{das2021confess}. Consequently, it is viable to further enhance the generalization capability of the FL model to local distributions (i.e., personalization) by selecting task-relevant features from $\bm{z}_i$ to construct sparse features.

\section{Experiment}\label{Experiments}
To demonstrate the effectiveness of FedPick, we conduct experiments on several cross-domain datasets and compare its results with those of several benchmark methods. The experimental details are discussed as follows.
\subsection{Dataset Description}\label{dataset_description}
\begin{figure}[h]
	\centering
	\includegraphics[width=3.0in]{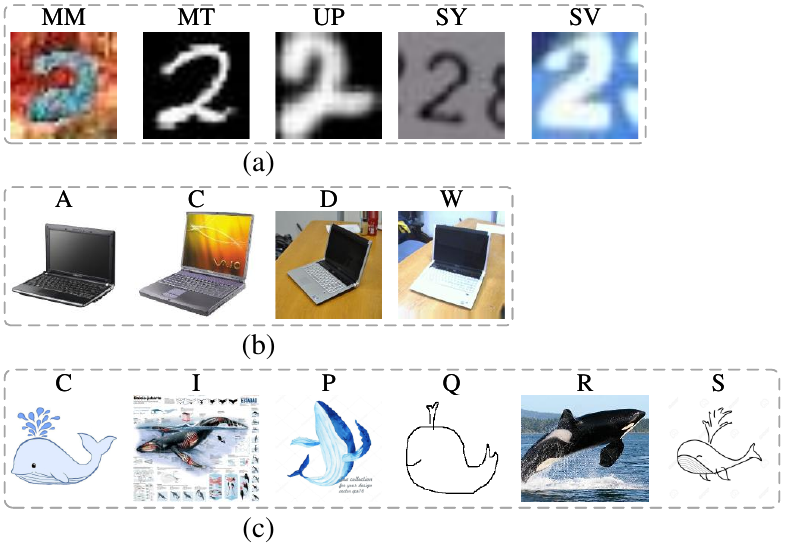}	\caption{Example images in different cross-domain datasets, (a) Digits-Five, (b) Office-Caltech-10, (c) DomainNet.}
        \label{example_images}
\end{figure}

Experiments are conducted on three commonly used cross-domain datasets: Digits-Five, Office-Caltech-10 and DomainNet. These datasets are all used for classification tasks. Fig. \ref{example_images} presents some example images in these datasets. The specific details of each dataset are provided below.

\textbf{Digits-Five.} The  Digits-Five dataset contains five datasets for handwritten character recognition with different background, namely MNIST-M (MM) \cite{SynthDigits}, MNIST (MT) \cite{MNIST}, USPS (UP) \cite{USPS}, SynthDigits (SY) \cite{SynthDigits}, and SVHN (SV) \cite{SVHN}. Each dataset consists of images from a single domain, categorized into $10$ classes. To construct the training dataset for each client, we sample $1000$ images from each dataset, resulting in a total of five clients participating in the training process. All images from the test datasets are utilized for evaluating the model. Prior to being fed into the model, all images are converted into RGB images of size $32 \times 32$.

\textbf{Office-Caltech-10.} The Office-Caltech-10 \cite{Office-Caltech} dataset is composed of images captured by cameras with diverse imaging parameters. It is categorized into four domains: Amazon (A), Caltech (C), DSLR (D), and Webcam (W). Each domain encompasses $10$ classes of images. For our experiments, we select $125$ training images from each domain and assign them to a single client. The test dataset is exclusively reserved for evaluation purposes. To standardize the input, the images are transformed into RGB format with dimensions of $256 \times 256$ pixels. Additionally, random flipping and rotation techniques are applied to augment the dataset.

\textbf{DomainNet.} The DomainNet \cite{DomainNet} dataset contains six domains: Clipart (C), Infograph (I), Painting (P), Quickdraw (Q), Real (R), and Sketch (S). These domains consist of images with various artistic styles, such as painting and sketching. Originally, each domain contains $345$ classes. But for our experiments, we select $10$ commonly used classes to construct our dataset. For each domain, we sample $500$ training images to create the training dataset for a single client. Similar to the Office-Caltech-10 dataset, all testing images are reserved exclusively for evaluation. The images are converted into $256 \times 256$ RGB images and augmented by randomly flipping and rotating before being fed into the model.

\subsection{Compared Methods}
To evaluate the effectiveness of FedPick, we perform a comparative analysis against the following methods:

\textbf{SingleSet:} This method trains and tests independent models for each client, utilizing only its own data. Despite not involving collaboration with other clients, it serves as a robust benchmark, especially as the quantity of local data increases.

\textbf{FedAvg:} FedAvg is a classical FL method that aggregates all local model parameters after several rounds of local updates on clients \cite{mcmahan2017communication}. 

\textbf{FedProx:} FedProx introduces a regularization term between the local and global models, which constrains the direction of local updates  \cite{li2020federated}. 

\textbf{FedPer:} This method utilizes a personalized classifier for each client, allowing adaptation of client-specific features extracted from the shared encoder \cite{FedPer}.

\textbf{FedRep:} Similar to FedPer, FedRep employs personalized classifiers for each client to adapt their features from a shared encoder \cite{collins2021exploiting}. However, it distinguishes itself by performing multiple local updates with respect to the personalized classifier for every update of the global encoder.

\textbf{LG-FedAvg:} In contrast to FedPer and FedRep, LG-FedAvg maintains personalized encoders on each client while sharing a global classifier across all clients \cite{liang2020think}.

\textbf{FedBN:} This method localizes BN layer parameters on each client while sharing the remaining parameters \cite{li2021fedbn}.

\subsection{Implementation Details}
For Digits-Five, we employ a CNN consisting of multiple convolutional and FC layers. For Office-Caltech-10 and DomainNet, we make use of a modified version of AlexNet \cite{krizhevsky2017imagenet}. This modification involves integrating a BN layer after each convolutional and FC layer.

The PFSM is plugged on the features generated by the global encoder. The feature dimension of Digits-Five is $8192$, while the feature dimensions of Office-Caltech-10 and DomainNet are both $4096$. Gumabel sampling is carried out using a compact FC network with the following architecture: [Linear($d$, $\frac{d}{2}$)] -ReLU - Linear($d$, $\frac{d}{2}$) - Gumbel Sigmoid], where $d$ represents the feature dimension.

\begin{table*}[!ht]
    \centering
    \caption{Test accuracy on Digits-Five, Office-Caltech-10 and DomainNet.}
    \begin{tabular}{l c c c c c | c c c c | c c c c c c}
    \bottomrule
         Method & \textbf{MM} & \textbf{MT} & \textbf{UP} & \textbf{SY} & \textbf{SV} & \textbf{A} & \textbf{C} & \textbf{D} & \textbf{W} & \textbf{C} & \textbf{I} & \textbf{P} & \textbf{Q} & \textbf{R} & \textbf{S} \\ 
    \hline
         SingleSet  & \makecell[c]{82.45 \\ (0.42)}  & \makecell[c]{95.89 \\ (0.17)}  & \makecell[c]{97.94 \\ (0.13)}  & \makecell[c]{84.21 \\ (0.17)}  & \makecell[c]{70.63 \\ (0.85)} & \makecell[c]{71.35 \\ (1.14)}  & \makecell[c]{47.91 \\ (0.44)} & \makecell[c]{98.75 \\ (1.53)} & \makecell[c]{93.22 \\ (1.07)} & \makecell[c]{64.83 \\ (0.21)} & \makecell[c]{35.28 \\ (0.56)} & \makecell[c]{53.86 \\ (0.36)} & \makecell[c]{82.44 \\ (0.33)} & \makecell[c]{69.20 \\ (0.45)} & \makecell[c]{56.39 \\ (1.01)} \\ \hline
         FedAvg & \makecell[c]{80.93 \\ (0.49)} & \makecell[c]{96.88 \\ (0.15)} & \makecell[c]{97.37 \\ (0.22)} & \makecell[c]{83.98 \\ (0.45)} & \makecell[c]{73.19 \\ (0.61)} & \makecell[c]{62.92 \\ (1.21)}  & \makecell[c]{53.69 \\ (0.99)} & \makecell[c]{82.50 \\ (2.50)} & \makecell[c]{94.58 \\ (1.27)} & \makecell[c]{71.06 \\ (0.72)} & \makecell[c]{33.24 \\ (0.98)} & \makecell[c]{56.16 \\ (0.74)} & \makecell[c]{68.24 \\ (1.18)} & \makecell[c]{63.78 \\ (0.56)} & \makecell[c]{60.76 \\ (1.24)} \\
         FedProx   & \makecell[c]{81.71 \\ (0.20)} & \makecell[c]{96.92 \\ (0.24)} & \makecell[c]{97.29 \\ (0.33)} & \makecell[c]{83.79 \\ (0.22)} & \makecell[c]{71.78 \\ (0.79)} & \makecell[c]{61.35 \\ (1.79)}  & \makecell[c]{53.78 \\ (0.74)} & \makecell[c]{82.50 \\ (3.19)} & \makecell[c]{95.59 \\ (1.34)} & \makecell[c]{70.61 \\ (0.80)} & \makecell[c]{32.88 \\ (0.10)} & \makecell[c]{55.22 \\ (1.25)} & \makecell[c]{67.02 \\ (0.96)} & \makecell[c]{63.47 \\ (0.52)} & \makecell[c]{58.16 \\ (1.28)} \\
         FedPer   & \makecell[c]{82.21 \\ (0.35)} & \makecell[c]{96.87 \\ (0.25)} & \makecell[c]{97.63 \\ (0.44)} & \makecell[c]{84.26 \\ (0.40)} & \makecell[c]{70.57 \\ (0.61)} & \makecell[c]{63.65 \\ (0.83)}  & \makecell[c]{54.31 \\ (0.52)} & \makecell[c]{83.75 \\ (3.64)} & \makecell[c]{94.58 \\ (1.27)} & \makecell[c]{68.90 \\ (0.26)} & \makecell[c]{35.34 \\ (0.85)} & \makecell[c]{54.25 \\ (0.62)} & \makecell[c]{75.28 \\ (0.89)} & \makecell[c]{66.77 \\ (0.77)} & \makecell[c]{54.84 \\ (0.52)} \\
         FedRep   & \makecell[c]{80.89 \\ (0.67)} & \makecell[c]{96.30 \\ (0.21)} & \makecell[c]{97.65 \\ (0.18)} & \makecell[c]{83.08 \\ (0.52)} & \makecell[c]{66.07 \\ (1.48)} & \makecell[c]{63.44 \\ (2.24)} & \makecell[c]{49.96 \\ (1.81)} & \makecell[c]{80.63 \\ (5.00)} & \makecell[c]{92.20 \\ (0.83)} & \makecell[c]{65.86 \\ (0.98)} & \makecell[c]{34.31 \\ (0.81)} & \makecell[c]{50.92 \\ (0.40)} & \makecell[c]{69.66 \\ (2.08)} & \makecell[c]{63.04 \\ (0.74)} & \makecell[c]{52.24 \\ (1.13)} \\
         LG-FedAvg   & \makecell[c]{82.42 \\ (0.23)} & \makecell[c]{95.87 \\ (0.05)} & \makecell[c]{97.78 \\ (0.09)} & \makecell[c]{84.34 \\ (0.42)} & \makecell[c]{70.84 \\ (0.49)} & \makecell[c]{71.88 \\ (0.57)} & \makecell[c]{52.36 \\ (1.10)} & \makecell[c]{96.88 \\ (1.98)} & \makecell[c]{97.97 \\ (0.68)} & \makecell[c]{69.77 \\ (0.67)} & \makecell[c]{35.83 \\ (0.80)} & \makecell[c]{59.39 \\ (1.15)} & \makecell[c]{\textbf{82.42} \\ \textbf{(0.45)}} & \makecell[c]{73.05 \\ (0.68)} & \makecell[c]{61.77 \\ (0.21)} \\
         FedBN      & \makecell[c]{83.53 \\ (0.26)} & \makecell[c]{97.24 \\ (0.08)} & \makecell[c]{98.45 \\ (0.09)} & \makecell[c]{85.58 \\ (0.31)} & \makecell[c]{78.22 \\ (0.30)} & \makecell[c]{72.40 \\ (0.87)}  & \makecell[c]{54.22 \\ (1.41)} & \makecell[c]{97.50 \\ (1.25)} & \makecell[c]{\textbf{98.64} \\ \textbf{(0.68)}} & \makecell[c]{71.22 \\ (1.00)} & \makecell[c]{34.82 \\ (0.42)} & \makecell[c]{59.55 \\ (1.02)} & \makecell[c]{80.80 \\ (0.39)} & \makecell[c]{70.29 \\ (0.56)} & \makecell[c]{63.03 \\ (0.63)} \\ \hline
         FedPick     & \makecell[c]{\textbf{89.60} \\ \textbf{(0.34)}} & \makecell[c]{\textbf{97.81} \\ \textbf{(0.02)}} & \makecell[c]{\textbf{98.81} \\ \textbf{(0.04)}} & \makecell[c]{\textbf{91.34} \\ \textbf{(0.25)}} & \makecell[c]{\textbf{82.87} \\ \textbf{(0.19)}} & \makecell[c]{\textbf{74.06} \\ \textbf{(1.06)}}  & \makecell[c]{\textbf{57.51} \\ \textbf{(0.36)}} & \makecell[c]{\textbf{100.00}\\ \textbf{(0.00)}} & \makecell[c]{ \textbf{98.64} \\ \textbf{(0.68)}} & \makecell[c]{\textbf{74.07} \\ \textbf{(0.49)}} & \makecell[c]{\textbf{37.05} \\ \textbf{(0.92)}} & \makecell[c]{\textbf{63.26} \\ \textbf{(0.82)}} & \makecell[c]{81.76 \\ 
        (0.42)} & \makecell[c]{\textbf{74.38} \\ \textbf{(0.50)}} & \makecell[c]{\textbf{66.75} \\ \textbf{(1.00)}} \\
    \toprule
    \end{tabular}
    \label{exp_all}
\end{table*}

\begin{figure*}[!ht]
	\centering
	\includegraphics[width=6.5in]{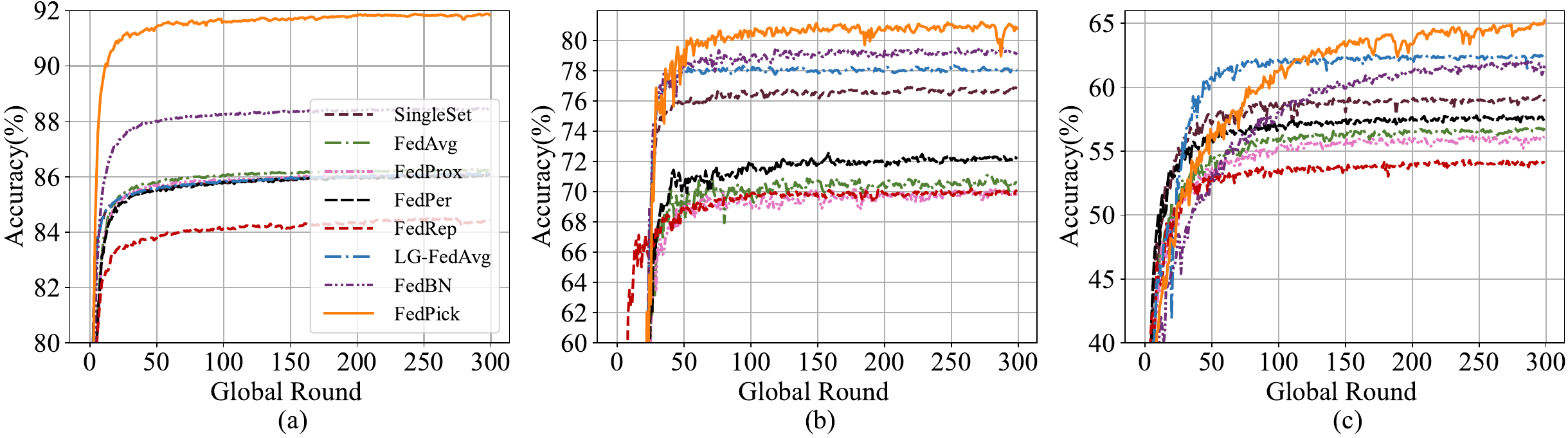}
	\caption{Variations in test accuracy throughout the training process: (a) Digits-Five, (b) Office-Caltech-10, (c) DomainNet.}
        \label{acc_curve}
\end{figure*}

All experiments are implemented using the PyTorch \cite{paszke2019pytorch} framework and executed on a four-card Nvidia V100 cluster. We use SGD with momentum to update the model. The learning rate and momentum are set to $0.01$ and $0.5$, respectively, in all experiments. Across all methods, a batch size of $64$ is employed during the local updating process. The local epoch is set to $1$ for all methods except for FedRep. In the case of FedRep, the local update consists of $5$ epochs, where the first $4$ epochs are dedicated to optimizing the classifier, and the final epoch focuses on optimizing the encoder. The total number of global communication rounds is set to $300$. To mitigate the randomness of the experimental results, we repeat all experiments five times. The mean and standard deviation values of the best test accuracy during the FL training are presented in the following sections. 

The hyperparameter $\mu$ adopted to balance the loss terms in FedProx is set to $0.01$ in the experiments. For LG-FedAvg, a pre-training phase comprising $20$ global rounds is performed by aggregating all parameters on Office-Caltech-10 and DomainNet. The hyperparameter combinations of FedPick are detailed in Table \ref{hyper-parameters}. 

\begin{table}[h]
    \centering
    \caption{Hyperparameters of FedPick.}
    \begin{tabular}{l c c c c}
    \toprule
          & $\tau$ & $\lambda_{lce}$ & $\lambda_{ent}$ & $\lambda_{dis}$ \\
    \midrule
         Digits-Five & 10.0 & 10.0 & 0.001 & 10.0 \\
         Office-Caltech-10 & 1.0 & 1.0 & 0.001 & 1.0 \\
         DomainNet & 1.0 & 1.0 & 0.001 & 1.0\\ \bottomrule
    \end{tabular}    
    \label{hyper-parameters}
\end{table}

\subsection{Experimental Results}
Tables \ref{exp_all} presents the basic experimental results on Digits-Five,  Office-Caltech-10, and DomainNet, respectively. These findings highlight that SingleSet, despite being trained on data from a single domain, performs remarkably well and serves as a strong benchmark method. However, traditional FL methods like FedAvg suffer from significant performance degradation due to the domain gaps between different clients, often failing to surpass the performance of SingleSet. On the other hand, FedPer, FedRep, LG-FedAvg, and FedBN achieve higher model accuracy by allowing the model to adapt to each domain through personalized parameters such as classifiers or BN layers in the encoder. Notably, our proposed method, FedPick, outperforms the compared methods on all datasets in most time, demonstrating its effectiveness in cross-domain FL. Fig. \ref{acc_curve} presents the variations in test accuracy during the training. FedPick can converge to higher accuracy stably compared with other FL methods.

\begin{table*}[h]
    \centering
    \caption{Ablation Study Results on Digits-Five, Office-Caltech-10 and DomainNet.}
    \begin{tabular}{l c c c c c | c c c c | c c c c c c}
    \bottomrule
         Setting & \textbf{MM} & \textbf{MT} & \textbf{UP} & \textbf{SY} & \textbf{SV} & \textbf{A} & \textbf{C} & \textbf{D} & \textbf{W} & \textbf{C} & \textbf{I} & \textbf{P} & \textbf{Q} & \textbf{R} & \textbf{S} \\ 
    \hline
         FedPick     & \makecell[c]{89.60 \\ (0.34)} & \makecell[c]{97.81 \\ (0.02)} & \makecell[c]{98.81 \\ (0.04)} & \makecell[c]{91.34 \\ (0.25)} & \makecell[c]{82.87 \\ (0.19)} & \makecell[c]{74.06 \\ (1.06)}  & \makecell[c]{57.51 \\ (0.36)} & \makecell[c]{100.00 \\ (0.00)} & \makecell[c]{98.64 \\ (0.68)} & \makecell[c]{74.07 \\ (0.49)} & \makecell[c]{37.05 \\ (0.92)} & \makecell[c]{63.26 \\ (0.82)} & \makecell[c]{81.76 \\ 
        (0.42)} & \makecell[c]{74.38 \\ (0.50)} & \makecell[c]{66.75 \\ (1.00)} \\ \hline
         w/o PFSM & \makecell[c]{83.53 \\ (0.26)} & \makecell[c]{97.24 \\ (0.08)} & \makecell[c]{98.45 \\ (0.09)} & \makecell[c]{85.58 \\ (0.31)} & \makecell[c]{78.22 \\ (0.30)} & \makecell[c]{72.40 \\ (0.87)}  & \makecell[c]{54.22 \\ (1.41)} & \makecell[c]{97.50 \\ (1.25)} & \makecell[c]{98.64 \\ (0.68)} & \makecell[c]{71.22 \\ (1.00)} & \makecell[c]{34.82 \\ (0.42)} & \makecell[c]{59.55 \\ (1.02)} & \makecell[c]{80.80 \\ (0.39)} & \makecell[c]{70.29 \\ (0.56)} & \makecell[c]{63.03 \\ (0.63)} \\
         w/o $L_{lce}$ & \makecell[c]{88.76 \\ (0.14)} & \makecell[c]{97.61 \\ (0.05)} & \makecell[c]{98.81 \\ (0.09)} & \makecell[c]{89.55 \\ (0.05)} & \makecell[c]{81.24 \\ (0.25)} & \makecell[c]{74.06 \\ (1.29)} & \makecell[c]{57.07 \\ (1.37)} & \makecell[c]{99.38 \\ (1.25)} & \makecell[c]{98.31 \\ (0.00)} & \makecell[c]{71.33 \\ (1.14)} & \makecell[c]{36.56 \\ (0.38)} & \makecell[c]{62.10 \\ (0.63)} & \makecell[c]{81.08 \\ (0.30)} & \makecell[c]{72.49 \\ (0.90)} & \makecell[c]{64.19 \\ (1.25)} \\
         w/o $L_{ent}$ & \makecell[c]{89.55 \\ (0.22)} & \makecell[c]{97.79 \\ (0.04)} & \makecell[c]{98.70 \\ (0.08)} & \makecell[c]{91.18 \\ (0.20)} & \makecell[c]{82.77 \\ (0.14)} & \makecell[c]{72.19 \\ (0.53)} & \makecell[c]{57.33 \\ (1.57)} & \makecell[c]{98.75 \\ (1.53)} & \makecell[c]{98.31 \\ (0.00)} & \makecell[c]{73.16 \\ (0.58)} & \makecell[c]{37.02 \\ (0.70)} & \makecell[c]{62.78 \\ (0.39)} & \makecell[c]{81.62 \\ (0.29)} & \makecell[c]{74.38 \\ (0.27)} & \makecell[c]{66.43 \\ (0.68)} \\
         w/o $L_{dis}$ & \makecell[c]{88.84 \\ (0.17)} & \makecell[c]{97.66 \\ (0.10)} & \makecell[c]{98.65 \\ (0.02)} & \makecell[c]{90.93 \\ (0.16)} & \makecell[c]{81.88 \\ (0.23)} & \makecell[c]{73.44 \\ (1.23)} & \makecell[c]{56.44 \\ (0.63)} & \makecell[c]{98.13 \\ (1.53)} & \makecell[c]{98.31 \\ (0.00)} & \makecell[c]{73.16 \\ (1.01)} & \makecell[c]{37.14 \\ (0.62)} & \makecell[c]{63.36 \\ (0.60)} & \makecell[c]{81.60 \\ (0.32)} & \makecell[c]{74.18 \\ (1.04)} & \makecell[c]{66.17 \\ (0.69)} \\
         Soft Mask & \makecell[c]{88.62 \\ (0.16)} & \makecell[c]{97.78 \\ (0.04)} & \makecell[c]{98.84 \\ (0.05)} & \makecell[c]{90.26 \\ (0.16)} & \makecell[c]{82.16 \\ (0.26)} & \makecell[c]{71.77 \\ (1.06)} & \makecell[c]{56.36 \\ (0.95)} & \makecell[c]{100.00 \\ (0.00)} & \makecell[c]{98.31 \\ (0.00)} & \makecell[c]{73.73 \\ (0.62)} & \makecell[c]{37.29 \\ (0.63)} & \makecell[c]{61.78 \\ (0.60)} & \makecell[c]{81.48 \\ (0.47)} & \makecell[c]{73.54 \\ (0.49)} & \makecell[c]{65.45 \\ (1.15)} \\
         Share BN & \makecell[c]{88.83 \\ (0.10)} & \makecell[c]{97.62 \\ (0.09)} & \makecell[c]{98.71 \\ (0.13)} & \makecell[c]{90.78 \\ (0.18)} & \makecell[c]{80.37 \\ (0.28)} & \makecell[c]{64.69 \\ (1.21)} & \makecell[c]{53.87 \\ (0.65)} & \makecell[c]{90.00 \\ (2.34)} & \makecell[c]{95.93 \\ (0.83)} & \makecell[c]{74.18 \\ (0.99)} & \makecell[c]{35.86 \\ (0.83)} & \makecell[c]{57.77 \\ (0.45)} & \makecell[c]{71.80 \\ (0.30)} & \makecell[c]{67.05 \\ (0.84)} & \makecell[c]{63.43 \\ (0.71)} \\
         Share PFSM & \makecell[c]{90.04 \\ (0.30)} & \makecell[c]{97.93 \\ (0.08)} & \makecell[c]{98.67 \\ (0.09)} & \makecell[c]{90.92 \\ (0.05)} & \makecell[c]{84.28 \\ (0.24)} & \makecell[c]{73.85 \\ (0.83)} & \makecell[c]{56.18 \\ (1.15)} & \makecell[c]{98.75 \\ (1.53)} & \makecell[c]{98.31 \\ (0.00)} & \makecell[c]{73.38 \\ (0.71)} & \makecell[c]{35.59 \\ (1.22)} & \makecell[c]{61.23 \\ (0.57)} & \makecell[c]{81.32 \\ (0.47)} & \makecell[c]{72.56 \\ (0.41)} & \makecell[c]{65.99 \\ (1.13)} \\
         w/o Ensemble & \makecell[c]{87.69 \\ (0.19)} & \makecell[c]{97.87 \\ (0.05)} & \makecell[c]{98.69 \\ (0.09)} & \makecell[c]{89.64 \\ (0.17)} & \makecell[c]{83.78 \\ (0.21)} & \makecell[c]{72.81 \\ (1.52)} & \makecell[c]{56.27 \\ (0.45)} & \makecell[c]{97.50 \\ (1.25)} & \makecell[c]{98.64 \\ (0.68)} & \makecell[c]{72.55 \\ (1.12)} & \makecell[c]{35.56 \\ (1.27)} & \makecell[c]{60.84 \\ (0.88)} & \makecell[c]{80.38 \\ (0.37)} & \makecell[c]{71.59 \\ (0.48)} & \makecell[c]{65.56 \\ (0.49)} \\
         
    \toprule
    \end{tabular}
    \label{ablation_all}
\end{table*}

\section{Additional Analysis}
In this section, we provide additional analysis of FedPick. First, we conduct several ablation studies to demonstrate the efficacy of the components employed in FedPick. Second, we delve into the feature analysis for FedPick, highlighting the effectiveness of feature selection operation. At last, we discuss the experiments conducted on the hyper-parameters employed in FedPick.
\subsection{Ablation Study}
In this subsection, we conduct ablation studies to illustrate the efficacy of the components introduced in FedPick. The results of the ablation study are presented in Table \ref{ablation_all}. In the majority of cases, when the components specifically designed for FedPick are removed, there is a noticeable decline in model performance, thereby substantiating the effectiveness of these components. The detailed experimental results are discussed as follows.

\textbf{Without PFSM.} FedPick leverages PFSM to select task-relevant features from the universal features. When PFSM is excluded from the FedPick framework, FedPick essentially reduces to FedBN. From the results presented in Table \ref{ablation_all}, it is evident that the removal of PFSM leads to a substantial decline in the performance of FedPick. This observation underscores the critical role of personalized feature selection in enhancing the FL model performance.

\textbf{Without $L_{lce}$, $L_{ent}$, and $L_{dis}$.} In Eq. (\ref{loss_total}), the loss terms $L_{lce}$, $L_{ent}$, and $L_{dis}$ are specifically designed to serve different purposes within the FedPick. These terms aim to minimize the classification loss of personalized features, maximize the entropy of predictions derived from task-irrelevant features, and facilitate knowledge distillation between global and personalized predictions, respectively. The purpose of this experiment is to evaluate the impact of these loss terms on the performance of FedPick.
The results presented in Table \ref{ablation_all} indicate that removing these loss terms leads to a drop in model accuracy, suggesting that these loss terms play a crucial role in enhancing the model performance of FedPick.

\textbf{Using Soft Mask.} In FedPick, we employ a hard binary mask, denoted as $\bm{m}$, to discretely select features (either selecting or not). However, an alternative approach involves employing a soft mask to re-weight the features in a continuous manner. In this experiment, we exclude the binary operation during the sampling and directly employ a soft mask, denoted as $\bm{m}_s$, to re-weight the features. The re-weighting process is shown in Eq. (\ref{soft_mask}),  where $\bm{z}^g$ represents the global features, $\bm{m}_s$ denotes the soft mask, $\bm{z}_s^p$ and $\bm{z}_s^u$ represents the task-relevant and task-irrelevant features, respectively. The results depicted in Table \ref{ablation_all} indicate that although the soft feature selection mechanism is effective in the context of FedPick, it falls short of surpassing the performance achieved by the hard feature selection mechanism adopted in this paper.
\begin{equation}
\label{soft_mask}
\bm{z}_s^p = \bm{z}^g \odot \bm{m}_s, \quad
\bm{z}_s^u = \bm{z}^g \odot (1 - \bm{m}_s).
\end{equation}

\textbf{Sharing BN layers.} BN is commonly employed by default in most state-of-the-art (SOTA) CNNs. Nevertheless, earlier studies have demonstrated that aggregating BN layers can lead to a notable performance degradation  \cite{li2021fedbn,andreux2020siloed}. Therefore, in FedPick, we adopt a personalized BN layer strategy to mitigate the mutual inference among multiple domains, following the approach taken by numerous previous studies \cite{li2021fedbn,andreux2020siloed,sun2021partialfed, mills2021multi,wang2023does}. In this experiment, we aim to demonstrate the effectiveness of personalizing the BN layers. As shown in Table \ref{ablation_all}, when the BN layers are shared across clients, there is a noticeable decrease in accuracy, consistent with the findings reported in prior research. Given the observed benefits of personalizing BN layers in a cross-domain scenario and the ease of integration into existing frameworks, we decide to incorporate personalized BN layers into FedPick, aiming to further enhance its performance.

\textbf{Sharing PFSM.} In this paper, the PFSM is localized on each client to specially select the personalized features based on each client's data distribution. This experiment aims to showcase the efficacy of personalizing the PFSM. The results presented in Table \ref{ablation_all} demonstrate that sharing the PFSM parameters leads to improved model performance for simpler datasets like Digits-Five. However, as the complexity of the dataset increases, as observed in Office-Caltech-10 and DomainNet, sharing the PFSM parameters negatively affects the model performance.

\textbf{Without Ensemble.} In FedPick, the final predictions are obtained by ensembling the predictions derived from both global features and personalized features. The purpose of this experiment is to showcase the effectiveness of this ensemble strategy. As depicted in Table \ref{ablation_all}, when using only a global classifier for prediction, there is a significant decrease in accuracy. Nevertheless, the accuracy still remains higher compared to FedBN. This higher accuracy can be attributed to the knowledge transfer from the personalized classifier to the global classifier through a process known as distillation.
\subsection{Analysis of Features}
\begin{figure}[h]
	\centering
	\includegraphics[width=3.3in]{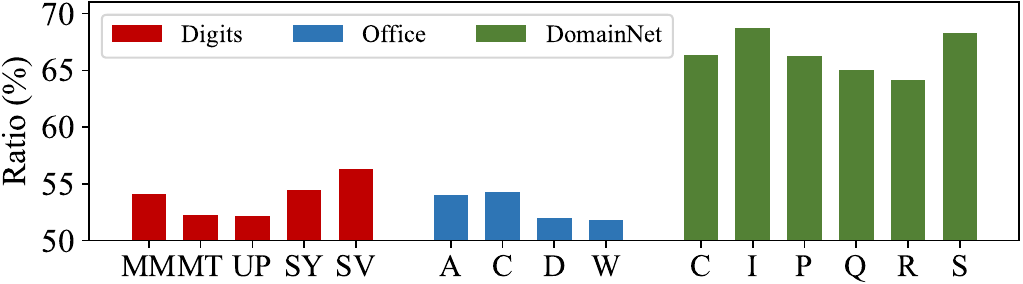}
	\caption{Feature selection ratios for different domains.}
        \label{mask_ratio}
\end{figure}

\textbf{Selection Ratios for Different Domains.}
Fig. \ref{mask_ratio} presents the ratios of selected task-relevant features to the total number of features for different domains. As the complexity of the dataset increases, the local task becomes more challenging, leading to a higher proportion of selected features to accomplish the task. For instance, the average selection ratio of DomainNet is significantly higher compared to the other two datasets. Similarly, within the same dataset, a more intricate domain tends to require a larger number of features. For instance, SVHN necessitates more features compared to other domains within the Digits-Five dataset. The complexity of the datasets can be discerned from the example images presented in Fig. \ref{example_images}. These findings demonstrate the adaptive feature selection capability of FedPick, allowing it to tailor the feature selection process based on the unique data distribution of each client.

\begin{figure}[h]
	\centering
	\includegraphics[width=3.3in]{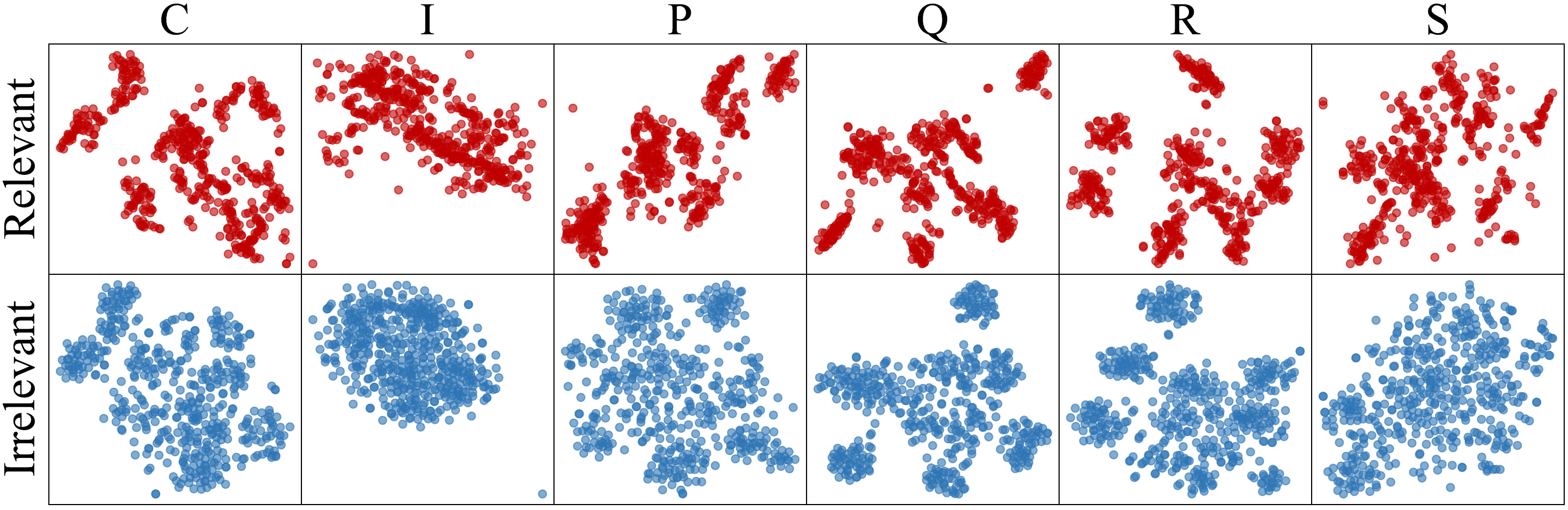}
	\caption{T-SNE \cite{van2008visualizing} visualization of task-relevant and task-irrelevant features on the DomainNet dataset.}
        \label{feature_visualization}
\end{figure}

\begin{figure}[h]
	\centering
	\includegraphics[width=3.3in]{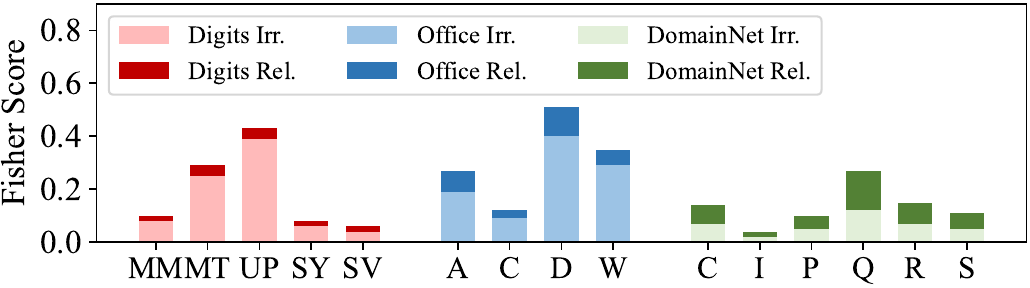}
	\caption{Fisher Scores of task-relevant and task-irrelevant features on different domains.}
        \label{feature_score}
\end{figure}

\textbf{Feature Discrimination.} Fig. \ref{feature_visualization} presents the T-SNE visualization \cite{van2008visualizing} of task-relevant and task-irrelevant features on the DomainNet dataset. The visualization demonstrates that task-relevant features exhibit higher discriminative characteristics, as they are more consistent within classes and scattered between classes compared to task-irrelevant features. This observation highlights the effective feature selection capability of FedPick in selecting important features for the local task. To provide a quantitative evaluation of feature discrimination, we further utilize the Fisher Score, as depicted in Fig. \ref{feature_score}. The figure reveals that task-relevant features possess higher Fisher Scores in comparison to task-irrelevant features, which aligns with the findings from Fig. \ref{feature_visualization}. Additionally, it is worth noting that the disparity in Fisher Scores between task-relevant and task-irrelevant features increases as the dataset complexity rises. This finding underscores the greater usefulness of feature selection for more complex datasets.

\begin{figure}[h]
	\centering
	\includegraphics[width=3.5in]{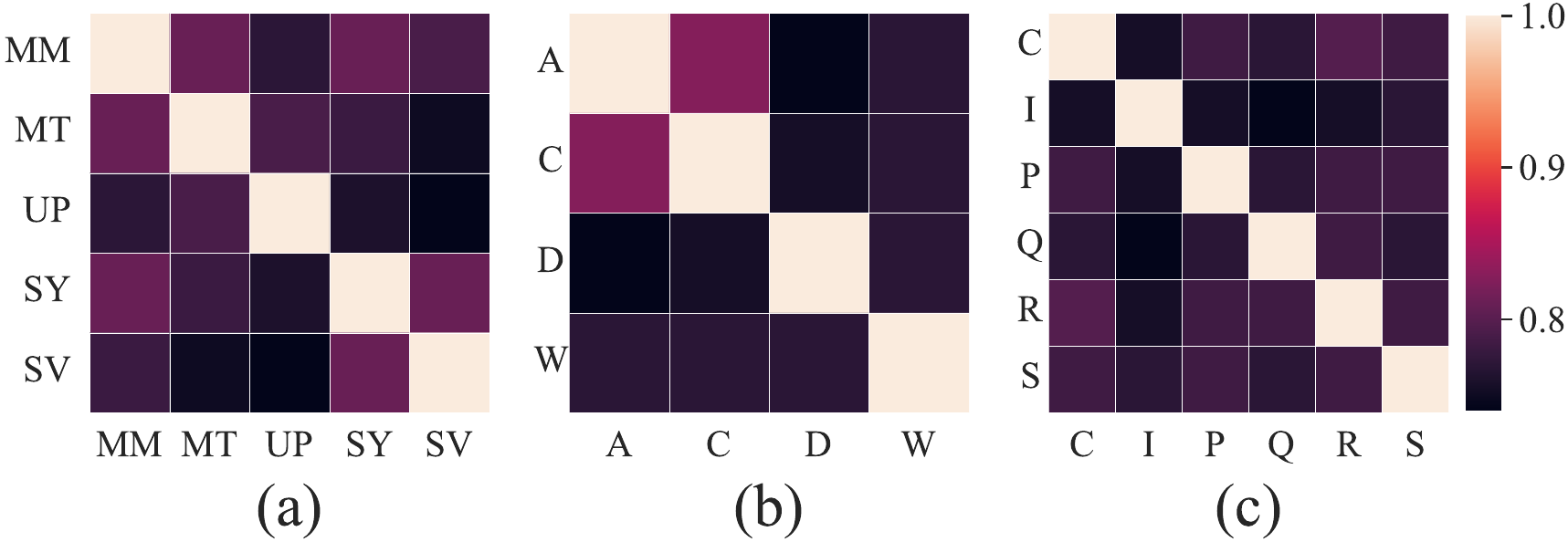}
	\caption{Overlapped ratios of important features among different domains, (a) Digits-Five, (b) Office-Caltech-10, (c) DomainNet.}
        \label{cover_ratio}
\end{figure}

\textbf{Overlapped Ratio of Important Features.}
In our analysis, we consider the top 50\% of the most frequently selected features on a test dataset as important features for the entire domain. To calculate the overlapped ratio of these important features between different domains, we determine how many of these features are common or shared across the domains under study. The overlapped ratio between two domains is defined as the ratio of the cardinal of the intersection to the union of important features. The overlapped ratios of important features between different domains are shown in Fig. \ref{cover_ratio}. It can be seen that there are a large proportion (about 20\% to 30\%) of important features that do not overlap between different domains, indicating that each domain possesses its own set of significant features. This result further supports the motivation behind FedPick. Additionally, within the same dataset, domains with a higher degree of similarity exhibit a greater overlapped ratio of important features. For example, MNIST and MNIST-M in the Digits-Five domain demonstrate a higher overlapped ratio, indicating that similar domains tend to share important features.
\subsection{Effect of Hyperparameters}

\begin{figure}[h]
	\centering
	\includegraphics[width=3.3in]{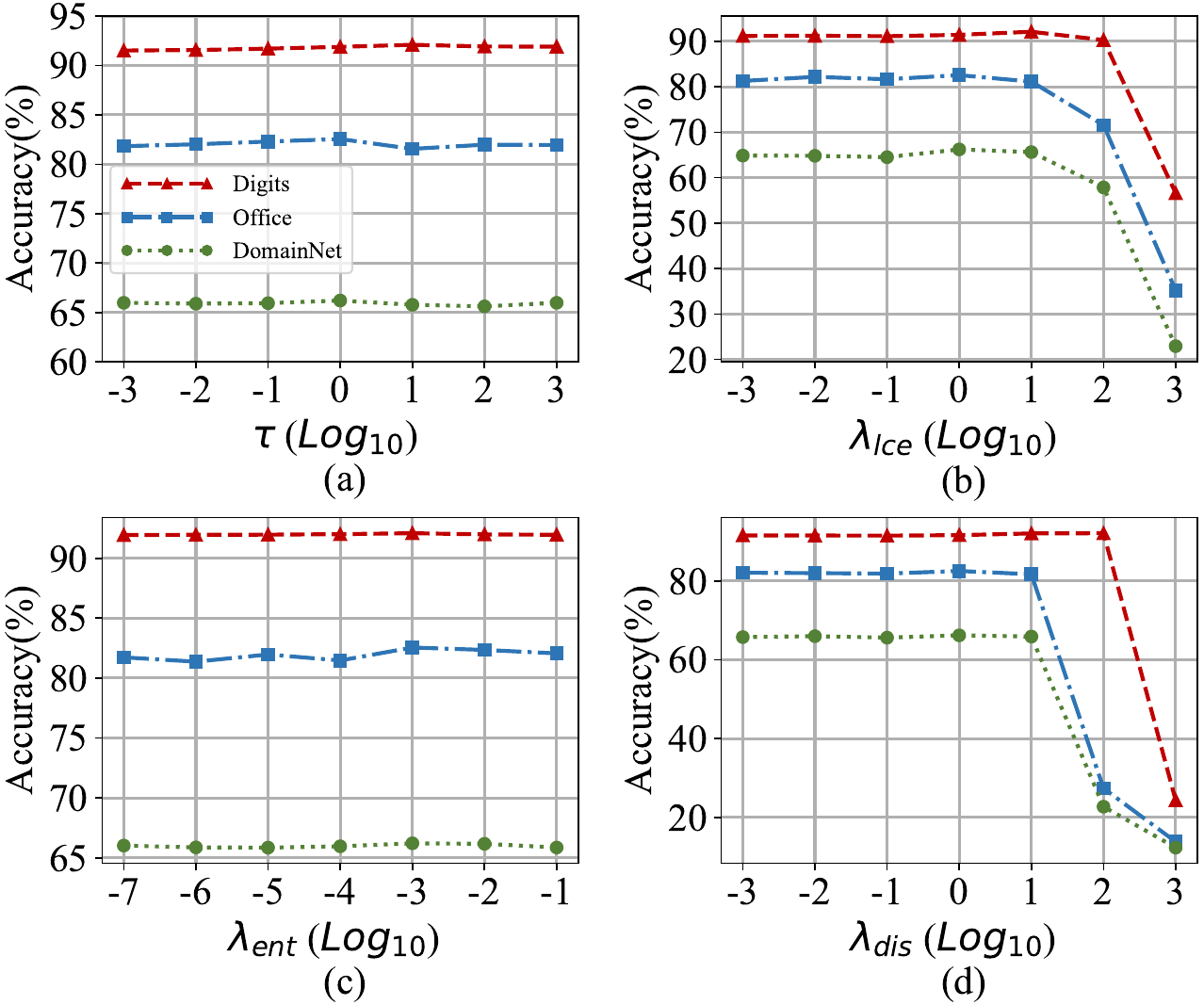}
	\caption{Test accuracy with different hyperparameters in Eq. (\ref{loss_total}): (a) $\tau$, (b) $\lambda_{lce}$, (c) $\lambda_{ent}$, (d) $\lambda_{dis}$.}
    \label{Lambda}
\end{figure}

\textbf{Hyperparameters in Loss Function.} The total loss function in Eq. (\ref{loss_total}) incorporates four hyperparameters, that is $\tau$, $\lambda_{lce}$, $\lambda_{ent}$, and $\lambda_{dis}$. The accuracy obtained with different hyperparameter settings is depicted in Fig. \ref{Lambda}. It can be observed that the model's performance remains stable with respect to changes in $\tau$. However, the performance is more sensitive to variations in $\lambda_{lce}$ and $\lambda_{dis}$. Although the curve representing $\lambda_{ent}$ in Figure \ref{loss_total} appears stable, we encountered gradient explosion issues when increasing its value during training. Consequently, in our experiments, we only report the results for smaller values of $\lambda_{ent}$ ranging from $10^{-7}$ to $10^{-1}$. 

\begin{figure}[!ht]
	\centering
	\includegraphics[width=3.5in]{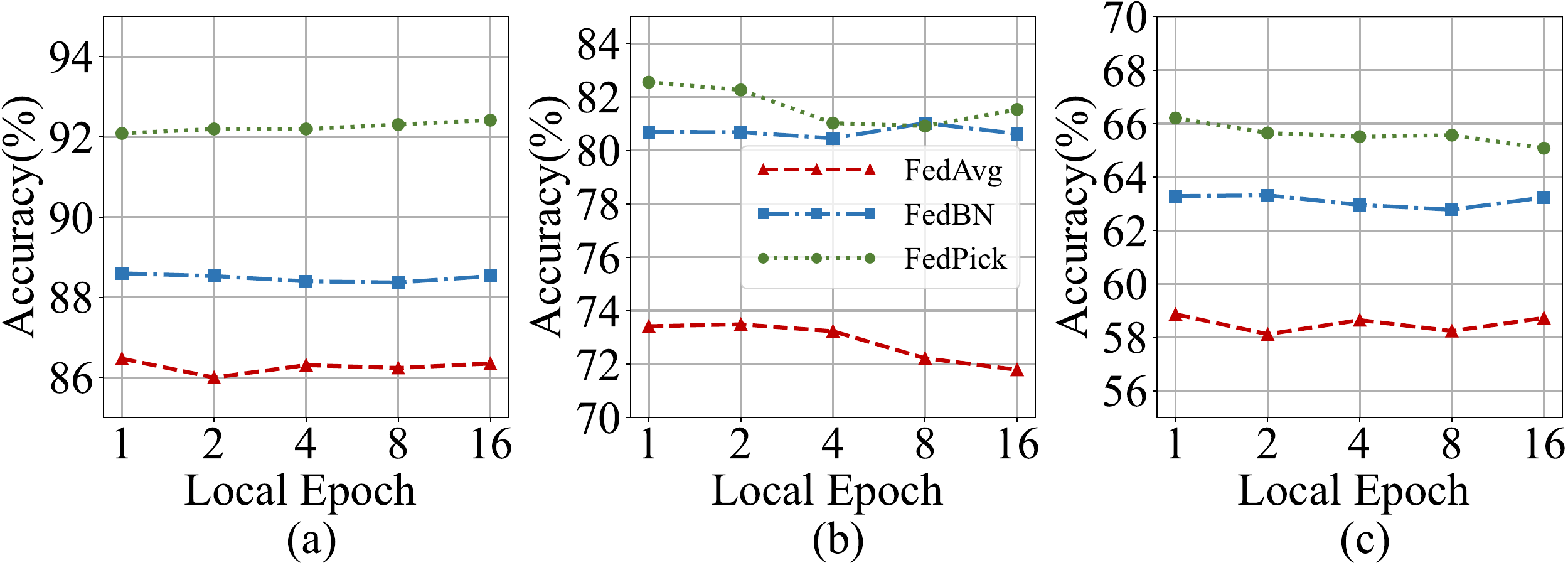}
	\caption{Test accuracy with different local epochs: (a) Digits-Five, (b) Office-Caltech-10, (c) DomainNet.}
        \label{local_epoch}
\end{figure}

\begin{figure}[!ht]
	\centering
	\includegraphics[width=3.5in]{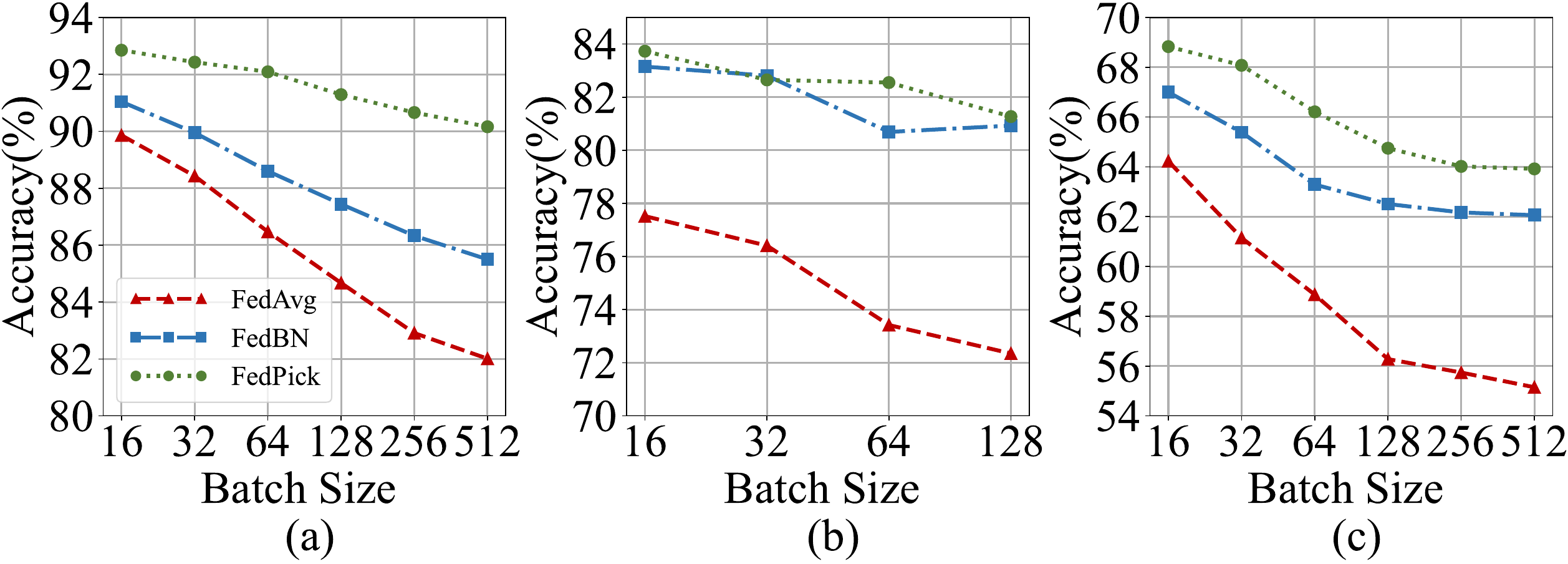}
	\caption{Test accuracy with different batch sizes: (a) Digits-Five, (b) Office-Caltech-10, (c) DomainNet.}
        \label{BS}
\end{figure}

\textbf{Hyperparameters for FL System.} We consider two hyperparameters that are important to FL, that are the local update epoch and local batch size. Fig. \ref{local_epoch} illustrates the model accuracy with varying local epochs. It can be observed that in cross-domain FL, the models of all FL methods exhibit a certain level of robustness to changes in local epochs. Particularly for simpler datasets like Digits-Five, increasing local epochs can even lead to a slight improvement in model accuracy. Fig. \ref{BS} displays the accuracy with different batch sizes. While the model accuracy decreases as the local batch size increases for all FL methods, FedPick demonstrates greater robustness to batch size variations and outperforms other FL methods across different batch sizes.
\section{Conclusion}
In this paper, we propose FedPick, a novel PFL framework for cross-domain scenario that works within the low-dimensional feature space. 
We are motivated by the observation that existing FL methods often extract redundant features, which can limit model performance. 
To address this issue, FedPick enhances model performance by adaptively selecting task-relevant features for each domain, leveraging its specific data distribution.
To achieve feature selection, FedPick introduces a PFSM for each client. The PFSM employs reparameterization techniques to make the discretized feature selection process differentiable. 
This enables easy integration of the PFSM into the backbone model, allowing it to be trained end-to-end using local data on each client.
Our experimental results demonstrate that FedPick successfully mitigates feature redundancy by selecting task-relevant features based on the data distribution of each client. 
Consequently, the proposed framework significantly improves model performance.
FedPick provides an effective approach to achieving PFL within a low-dimensional feature space. This not only simplifies implementation but also enhances interpretability, thereby showcasing its potential for practical applications in cross-domain FL.
\section*{Acknowledgment}
This work was supported in part by National Natural Science Foundation of China under Grant No. 61976012. 
 
\bibliographystyle{IEEEtran}
\bibliography{reference}

\end{document}